\documentclass{article}
\usepackage[preprint]{colm2026_conference}

\usepackage[T1]{fontenc}
\usepackage{microtype}
\usepackage{hyperref}
\usepackage{url}
\usepackage{booktabs}
\usepackage{amsmath}
\usepackage{enumitem}
\usepackage{tabularx}
\usepackage{lineno}
\usepackage{graphicx}
\usepackage{adjustbox}
\usepackage{multirow}
\usepackage{wrapfig}
\usepackage[breakable,skins]{tcolorbox}
\usepackage{listings}
\usepackage{caption}
\usepackage{float}
\usepackage{titletoc}
\usepackage{pifont}
\usepackage{wasysym}
\definecolor{cmarkgreen}{RGB}{34,139,67}
\definecolor{xmarkred}{RGB}{200,42,42}
\definecolor{pmarkamber}{RGB}{217,144,30}
\newcommand{\cmark}{\textcolor{cmarkgreen}{\ding{51}}}
\newcommand{\xmark}{\textcolor{xmarkred}{\ding{55}}}
\newcommand{\pmark}{\textcolor{pmarkamber}{\raisebox{0.15ex}{\scalebox{0.95}{\LEFTcircle}}}}

\definecolor{darkblue}{rgb}{0, 0, 0.5}
\hypersetup{colorlinks=true, citecolor=darkblue, linkcolor=darkblue, urlcolor=darkblue}

\title{%
  \adjustimage{height=1.5\baselineskip, valign=c, raise=2.1pt, margin=0 5pt 5pt 0}{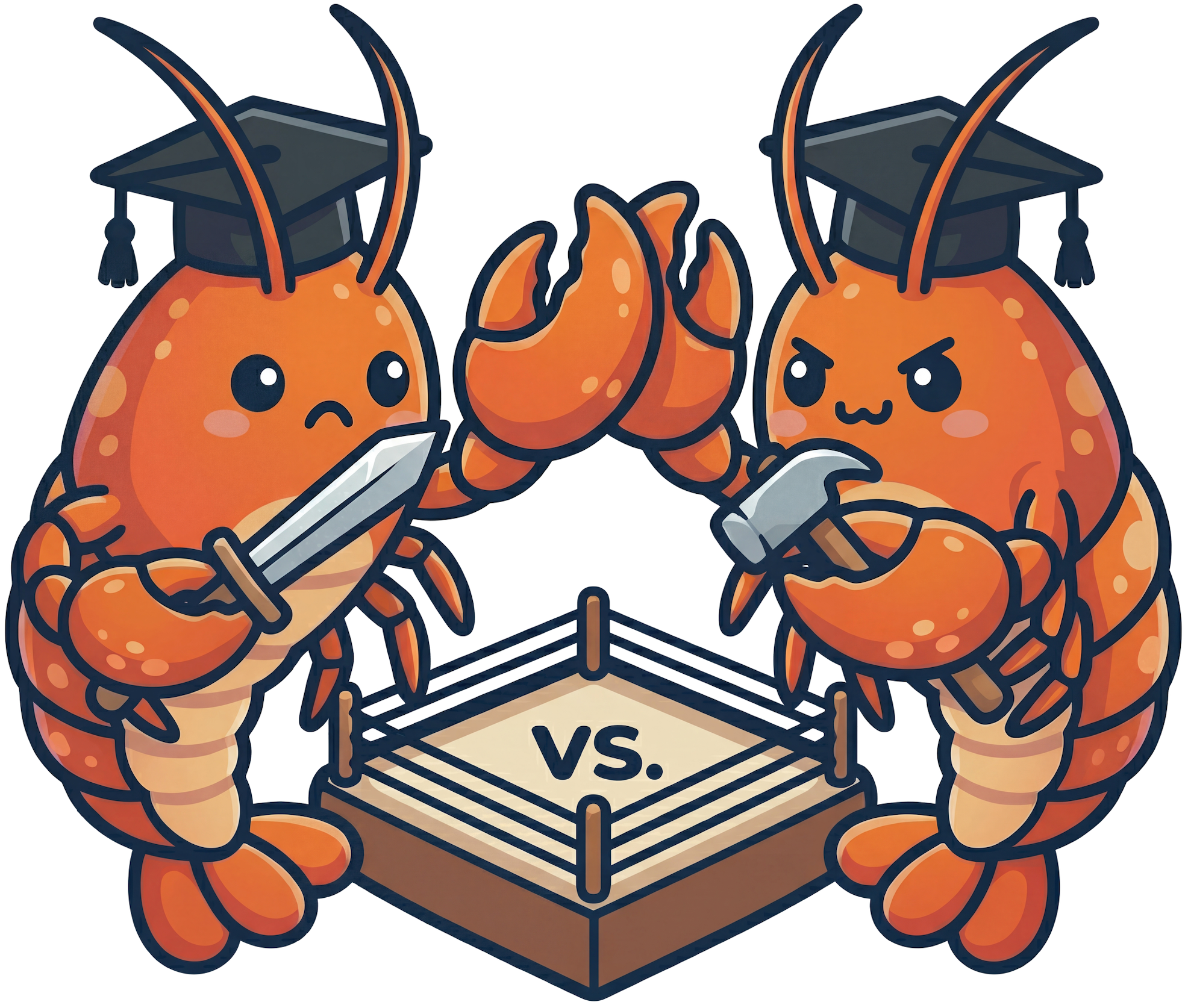}%
  ClawArena: Benchmarking AI Agents in Evolving\\ \vspace{-0.5em} Information Environments
}

\author{\textbf{Haonian Ji}$^{1}$\thanks{Equal contribution.}, 
\textbf{Kaiwen Xiong}$^{1*}$, 
\textbf{Siwei Han}$^1$, 
\textbf{Peng Xia}$^1$, 
\textbf{Shi Qiu}$^{1}$, 
\textbf{Yiyang Zhou}$^{1}$,\\
\textbf{Jiaqi Liu}$^{1}$, 
\textbf{Jinlong Li}$^{1}$, 
\textbf{Bingzhou Li}$^{1}$, 
\textbf{Zeyu Zheng}$^{3}$, 
\textbf{Cihang Xie}$^{2}$, 
\textbf{Huaxiu Yao}$^{1}$
\\
$^1$UNC-Chapel Hill\quad
$^2$University of California, Santa Cruz\quad \\
$^3$University of California, Berkeley \\
\texttt{\{haonianj, xkaiwen, huaxiu\}@cs.unc.edu}
}

\begin{document}

% Line numbers disabled for arXiv preprint

\maketitle

\vspace{-0.5em}
\begin{abstract}
\vspace{-0.5em}

AI agents deployed as persistent assistants must maintain correct beliefs as their information environment evolves. In practice, evidence is scattered across heterogeneous sources that often contradict one another, new information can invalidate earlier conclusions, and user preferences surface through corrections rather than explicit instructions. Existing benchmarks largely assume static, single-authority settings and do not evaluate whether agents can keep up with this complexity. We introduce \textsc{ClawArena}, a benchmark for evaluating AI agents in evolving information environments. Each scenario maintains a complete hidden ground truth while exposing the agent only to noisy, partial, and sometimes contradictory traces across multi-channel sessions, workspace files, and staged updates. Evaluation is organized around three coupled challenges: multi-source conflict reasoning, dynamic belief revision, and implicit personalization, whose interactions yield a 14-category question taxonomy. Two question formats, multi-choice (set-selection) and shell-based executable checks, test both reasoning and workspace grounding. \textsc{ClawArena} comprises 12 multi-turn scenarios spanning 337 evaluation rounds with 45 dynamic updates, evaluated across five agent frameworks and 18 language models from proprietary, community-accessible, and self-hosted sources. To capture behavioral consistency beyond correctness, we introduce the Composite Reliability Score (CRS), which combines Task Completion Rate with a Robustness term that rewards sustained success streaks and penalizes clustered failures. Experiments show that model capability accounts for a 29-point CRS range across models while framework design accounts for up to a 24-point range, that MetaClaw's skill overlay reliably improves Robustness without degrading accuracy, and that belief revision difficulty is determined by update design strategy rather than update volume. Code is available at \href{https://github.com/aiming-lab/ClawArena}{https://github.com/aiming-lab/ClawArena}.
\end{abstract}

\vspace{-0.5em}
\section{Introduction}
\vspace{-0.5em}

\begin{figure*}[t]
\centering
\includegraphics[width=\textwidth]{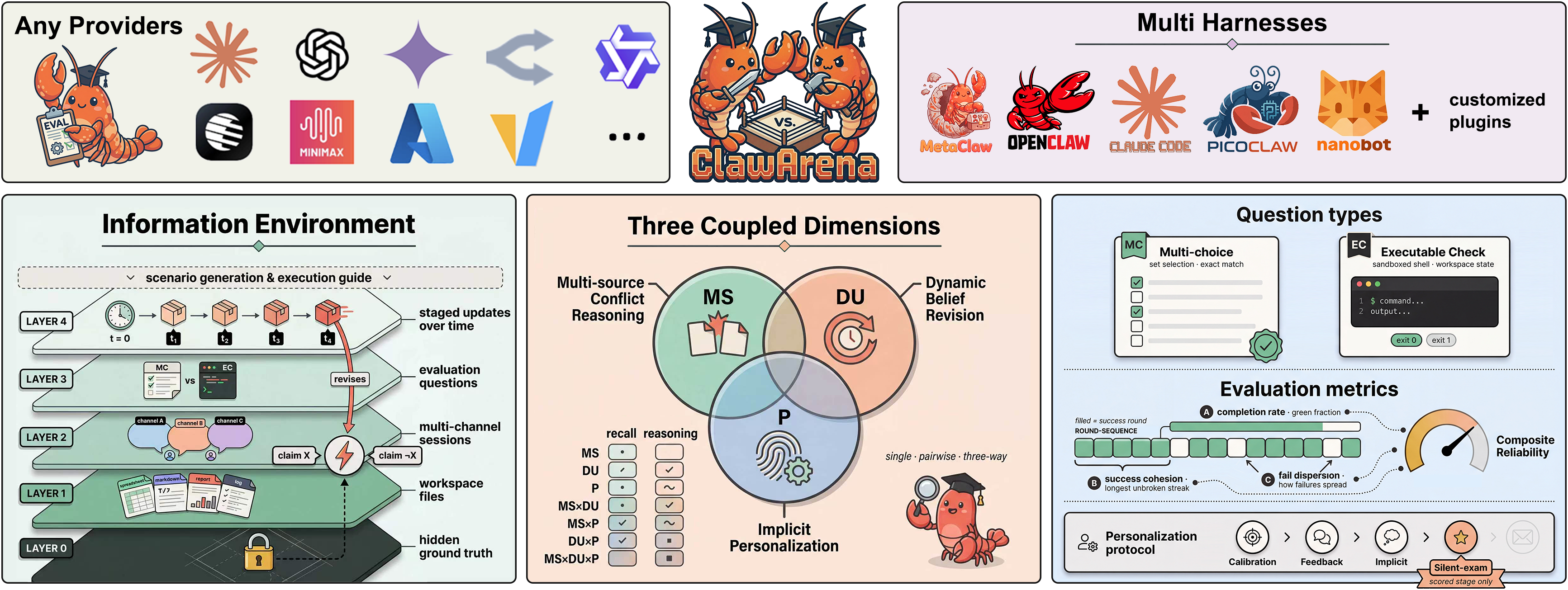}
\caption{Overview of \textsc{ClawArena}. An evolving information environment with conflicting and progressively updated evidence is structured along three coupled evaluation dimensions, instantiated through two complementary question formats, and summarized by a composite reliability score that jointly reflects accuracy and behavioural consistency.}
\label{fig:overview}
\vspace{-1.5em}
\end{figure*}

Maintaining a correct view of ongoing work is critical for AI agents deployed as persistent assistants, yet real information environments make this difficult: evidence is scattered across heterogeneous sources (chat histories, workspace files, monitoring logs) that often contradict one another, for instance when multiple sources report conflicting timelines for the same incident. These environments create three coupled challenges. The agent must judge source reliability rather than naively aggregate (\emph{multi-source conflict reasoning}). New evidence arrives over time and can invalidate previously correct conclusions, so the agent must revise rather than accumulate (\emph{dynamic belief revision}). User preferences are rarely stated explicitly and surface through corrections and interaction patterns, so the agent must learn and apply them without reminders (\emph{implicit personalization}). An agent that handles conflicts but fails to revise, or revises but ignores format preferences, still produces unreliable output.

Existing evaluations test fragments of this problem but not the full setting. Task-oriented agent benchmarks typically provide a single authoritative environment \citep{jimenez2024swebench,liu2024agentbench,zhou2024webarena,xie2024osworld,mialon2024gaia}. Long-context and multi-hop QA benchmarks test retrieval and composition over static evidence \citep{yang2018hotpotqa,trivedi2022musique,bai2024longbench,hsieh2024ruler}. Memory benchmarks emphasize long-horizon recall but usually without explicit source conflict or silent preference retention \citep{maharana2024locomo,zhang2018personachat}. Taken together, existing benchmarks assume a static, single-authority information environment, leaving open the question of whether agents can maintain correct and up-to-date beliefs as their information environment evolves.

We introduce \textsc{ClawArena}, a benchmark for AI agents in evolving information environments (Figure~\ref{fig:overview}). Each scenario combines multi-channel session histories, workspace files, staged update packages, and a four-stage personalization protocol ending in silent-exam rounds with no preference reminders. Each scenario maintains a complete hidden ground truth, while the agent observes only noisy, partial, sometimes contradictory traces; correctness is verified against the ground truth rather than any single observable source. Observable materials are grounded in real-world empirical distributions covering message timing, contact frequency, and information overload. The three dimensions and their interactions yield a 14-category taxonomy that blocks one-dimensional scoring, and two question formats (multi-choice set-selection and shell-based executable checks) test reasoning and workspace grounding. The benchmark comprises 12 multi-turn scenarios spanning 337 evaluation rounds with 45 dynamic updates, used in full for all reported experiments.

Beyond the benchmark itself we contribute: (i) a reproducible three-stage construction pipeline with real-world empirical grounding; (ii) a 14-category diagnostic taxonomy enabling per-dimension and per-interaction failure analysis; (iii) the Composite Reliability Score (CRS), with a Robustness decomposition into Success Cohesion and Failure Dispersion, capturing behavioral consistency beyond raw accuracy; and (iv) experiments across five frameworks and 18 language models from proprietary, community-accessible, and self-hosted sources, showing that model capability dominates framework design in CRS range, that MetaClaw's skill overlay improves Robustness without degrading accuracy, and that belief revision difficulty is governed by update design rather than update volume.

\section{Related Work}
\label{sec:relatedwork}

\paragraph{Agent Benchmarks.}
Task-oriented benchmarks such as SWE-bench \citep{jimenez2024swebench}, AgentBench \citep{liu2024agentbench}, WebArena \citep{zhou2024webarena}, OSWorld \citep{xie2024osworld}, and GAIA \citep{mialon2024gaia} evaluate tool use, planning, and execution in largely single-authority environments and do not test adjudication across conflicting sources. Long-context and multi-hop QA benchmarks such as HotpotQA \citep{yang2018hotpotqa}, MuSiQue \citep{trivedi2022musique}, LongBench \citep{bai2024longbench}, and RULER \citep{hsieh2024ruler} stress retrieval and composition over static evidence; ConflictQA \citep{xie2024conflictqa} adds conflicting claims, but evidence remains fixed at inference time, so none evaluates revision after new evidence arrives. Memory and persona benchmarks such as LoCoMo \citep{maharana2024locomo} and PersonaChat \citep{zhang2018personachat} capture long-horizon recall and user-model consistency but do not jointly require cross-source conflict resolution, dynamic updates, and silent preference retention. ClawArena targets this joint setting.

\paragraph{Harness-native Benchmarks.}
A second emerging line of work evaluates models \emph{inside concrete agent harnesses} (OpenClaw, Claude Code, Codex, Gemini CLI, etc.) rather than as standalone reasoners. These include browser harnesses (\textsc{ClawBench}~\citep{zhang2026clawbench}), real or sandboxed productivity stacks (\textsc{PinchBench}~\citep{pinchbench2026}, \textsc{QwenClawBench}~\citep{qwenteam2026qwenclawbench}, \textsc{WildClawBench}~\citep{ding2026wildclawbench}, \textsc{ZClawBench}~\citep{zai2026zclawbench}), mock-service suites (\textsc{ClawsBench}~\citep{li2026clawsbench}, \textsc{Claw-Eval}~\citep{ye2026claweval}, \textsc{Claw-Eval-Live}~\citep{li2026clawevallive}), workspace simulations (\textsc{MetaClawBench}~\citep{xia2026metaclaw}), and multi-day coworker scenarios (\textsc{ClawMark}~\citep{meng2026clawmark}). Table~\ref{tab:related_compare} contrasts them across the four design axes that drive evolving information environments, together with task sourcing, execution mode, verification method, and the number of frameworks evaluated. Most prior work covers at most two of these axes; the closest entry, \textsc{ClawMark}, satisfies three but omits implicit personalization. \textsc{ClawArena} additionally instantiates a four-stage silent-exam preference protocol and reports across five distinct frameworks rather than one, making it the only benchmark to satisfy all four design axes simultaneously.

\begin{table}[t]
\centering
\small
\caption{Comparison with harness-native agent benchmarks. \cmark = supported, \xmark = unsupported, \pmark = partial (mechanism present but weakened, e.g., third-party messages exposed only as plain files rather than channel-tagged sessions, or preferences encoded as static persona rather than learned in silent rounds). MSC: agent must analyse user--third-party conversations distinguished by channel. DU: environment is overwritten between user turns (intra-loop tool-return dynamics do not count). MU: user re-engages with new queries across rounds. Pref.: implicit user preferences applied in silent-exam rounds.}
\label{tab:related_compare}
\resizebox{\linewidth}{!}{%
\begin{tabular}{@{}lllccccll l@{}}
\toprule
\textbf{Benchmark} & \textbf{Task sourcing} & \textbf{Exec. mode} & \textbf{MSC} & \textbf{DU} & \textbf{MU} & \textbf{Pref.} & \textbf{Verification} & \textbf{Frmw.} & \textbf{Scale (Q\,/\,Scen.)} \\
\midrule
ClawBench         & Manual pool                 & Live web              & \xmark & \xmark & \xmark & \xmark & rule+llm & 8 & 283\,/\,144 sites \\
Claw-Eval         & Curated from upstream       & Sandbox + mock        & \xmark & \xmark & \cmark & \xmark & rule+llm & 1 & 300\,/\,9 cats \\
Claw-Eval-Live    & Live signals (quarterly)    & Mock services         & \xmark & \xmark & \xmark & \xmark & rule+llm & 1 & 105\,/\,17 fam. \\
ClawMark          & Manual + AI synthesis       & Sandboxed services    & \cmark & \cmark & \cmark & \xmark & rule-based & 1 & 100\,/\,13 scen. \\
ClawsBench        & Expert-designed             & Mock services         & \cmark & \xmark & \xmark & \xmark & rule-based & 4 & 44\,/\,5 svc. \\
MetaClawBench     & Synthetic                   & Workspace sim         & \pmark & \cmark & \cmark & \pmark & rule-based & 1 & 346\,/\,30 days \\
PinchBench        & Manual (real-world)         & Real (OpenClaw)       & \xmark & \xmark & \xmark & \xmark & rule+llm & 1 & 23\,/\,8 cats \\
QwenClawBench     & Empirical (claimed)         & Real (Docker)         & \xmark & \xmark & \xmark & \pmark & rule+llm & 1 & 100\,/\,8 dom. \\
WildClawBench     & Manual (in the wild)        & Real (OpenClaw)       & \cmark & \xmark & \xmark & \pmark & rule+llm & 1 & 60\,/\,6 cats \\
ZClawBench        & Manual + synthetic          & Real + part. mock     & \xmark & \xmark & \xmark & \xmark & rule+llm & 1 & 116\,/\,6 cats \\
\midrule
\textbf{ClawArena (Ours)} & \textbf{Empirical synthesis} & \textbf{Multi-channel sim} & \cmark & \cmark & \cmark & \cmark & rule-based & \textbf{5} & \textbf{337\,/\,12 scen.} \\
\bottomrule
\end{tabular}%
}
\vspace{-1em}
\end{table}

\noindent \textbf{LLM Agents.}
Modern LLMs enable agents beyond single-turn QA~\citep{comanici2025gemini,openai2025o3,yao2022react,shinn2023reflexion}; reasoning-focused models handle complex multi-step tasks~\citep{zhao2024expel,tang2025agent,ouyang2025reasoningbank,chhikara2025mem0,liu2026simplemem,xia2026skillrl,xia2026metaclaw,omnisimplemem2026}; tool-augmented agents interact with APIs, interpreters, and file systems~\citep{team2025tongyi,qintoolllm,feng2025retool,xia2025agent0,liu2025agent0,jin2025search}; and recent work explores multi-agent coordination \citep{wu2024autogen,li2023camel}, domain-specific systems \citep{ji2025eduvisagent,han2026paper2figure}, and long-horizon planning. Most such architectures are evaluated in episodic settings that reset between tasks. ClawArena instead targets the persistent-assistant regime where beliefs accumulate across sessions and must be revised as the environment evolves.

\section{ClawArena}
\label{sec:clawarena}

\subsection{Overview}

Each ClawArena scenario simulates a realistic information environment an AI agent must navigate. A scenario comprises multi-channel session histories (5--7 channels such as Slack, email, and WeChat, 200--400 messages), workspace files (4--8 documents including monitoring logs, sprint notes, audit reports), staged update packages that inject new evidence over time, and a four-stage personalization protocol ending in silent-exam rounds. Each scenario maintains a complete hidden ground truth against which answers are verified, while the agent observes only noisy, partial, sometimes contradictory traces.

For example, a startup outage scenario presents conflicting recovery timelines across a direct message, monitoring export, and sprint notes; the agent must synthesize conflicts, revise as later updates arrive, and present results in the user's preferred format.

\textsc{ClawArena} comprises 12 scenarios spanning diverse professional contexts (retail analytics, finance, healthcare, information security, HR, education, research integrity, and others) with 337 evaluation rounds and 45 dynamic updates in total; representative scenarios are catalogued in Appendix~\ref{app:domain_gallery} and the full statistical breakdown is reported in Appendix~\ref{app:core_stats}. Two formats test complementary capabilities: multi-choice (set-selection) over 7--9 candidate statements, and shell-based executable checks verifying whether claims about workspace evidence hold in the actual files.

\begin{table}[ht]
\centering
\small
\caption{Question taxonomy of \textsc{ClawArena}. Each row shows a non-empty subset of \{MS, DU, P\} and a representative task.}
\label{tab:taxonomy}
\begin{tabularx}{\columnwidth}{@{}cl>{\raggedright\arraybackslash}X@{}}
\toprule
\textbf{Tag} & \textbf{Dimensions} & \textbf{Representative question} \\
\midrule
\multicolumn{3}{@{}l}{\textit{Single dimension}} \\[2pt]
MS & Multi-Source & Identify which statements are supported across a direct message, a group chat, and a log export. \\[3pt]
DU & Dynamic Update & Update the conclusion after a new audit file or appended message arrives. \\[3pt]
P  & Personalization & Recall or apply the learned output style without an explicit reminder. \\[2pt]
\midrule
\multicolumn{3}{@{}l}{\textit{Pairwise interaction}} \\[2pt]
MS{\small$\times$}DU & MS $\times$ DU & Reassess a cross-source claim after the evidence state changes. \\[3pt]
MS{\small$\times$}P  & MS $\times$ P  & Resolve source conflict while using the learned reporting style. \\[3pt]
DU{\small$\times$}P  & DU $\times$ P  & Apply preferences to a task that emerges only after an update. \\[2pt]
\midrule
\multicolumn{3}{@{}l}{\textit{Three-way interaction}} \\[2pt]
All & MS $\times$ DU $\times$ P & Synthesize updated cross-source evidence and present it in the learned style. \\
\bottomrule
\end{tabularx}
\vspace{-1.5em}
\end{table}

\subsection{Evaluation dimensions and taxonomy}
\label{sec:dimensions}

A persistent assistant's information environment has three structural traits causing distinct failure modes: information spans multiple conflicting sources, the evidence base evolves, and user expectations form from prior interactions rather than upfront statements. ClawArena frames evaluation around three matching dimensions, each overlooked by existing benchmarks:

\noindent \textbf{Multi-source conflict reasoning (MS).} Evidence is distributed across heterogeneous sources that may contradict one another (e.g., a DM claiming a four-minute outage versus a monitoring log recording 47 minutes). The agent must judge source reliability; accuracy is measured as exact-set match on cross-source conflict questions.

\noindent \textbf{Dynamic belief revision (DU).} New evidence can invalidate previously correct conclusions (e.g., a later audit reveals an incomplete fix). The agent must explicitly revise; the revision rate measures how often it updates its answer after contradictory evidence arrives.

\noindent \textbf{Implicit personalization (P).} User preferences emerge via corrections and interaction patterns (e.g., users reformat bullet outputs into tables). The agent must learn and apply these preferences without reminders; compliance is scored only in silent-exam rounds.

The three dimensions and their pairwise and three-way interactions yield seven combination categories (Table~\ref{tab:taxonomy}), each split into \emph{recall} (can the agent retrieve the evidence?) and \emph{reasoning} (can it draw the correct conclusion?) variants for 14 categories. This blocks scoring well by solving one dimension in isolation.

\subsection{Scenario design}
\label{sec:scenario}

\begin{wrapfigure}{r}{0.5\columnwidth}
\centering
\vspace{-1.5em}
\includegraphics[width=0.48\columnwidth]{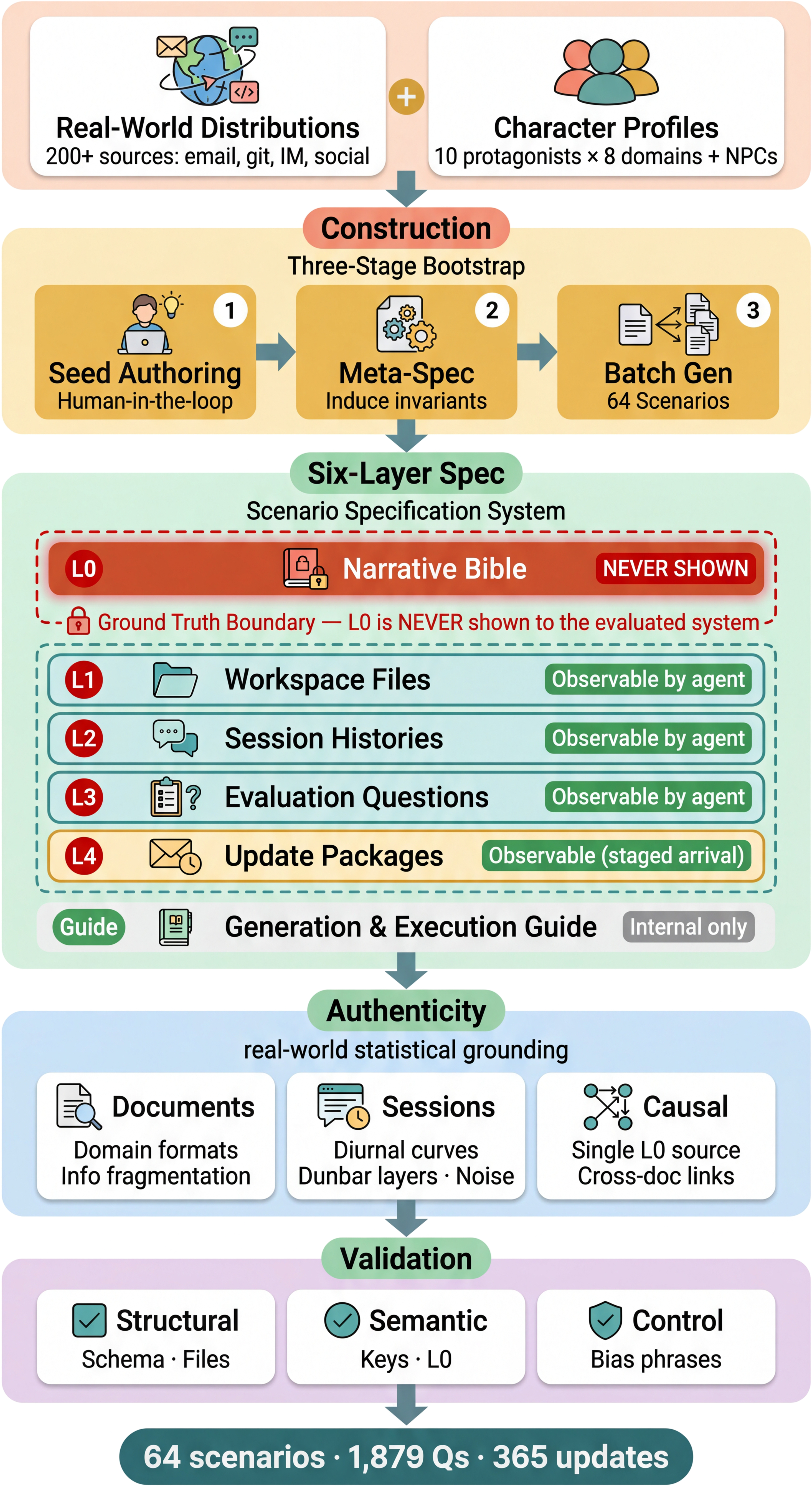}
\caption{\textsc{ClawArena} construction pipeline. Real-world distributions and character profiles feed a three-stage bootstrap, producing the 12-scenario benchmark organized into six layers and refined through three validation passes.}
\label{fig:pipeline}
\vspace{-2em}
\end{wrapfigure}

Each scenario is a six-layer specification. Layer~0 holds the hidden ground truth (objective timeline, contradiction map, answer provenance). Layers~1--4 are agent-visible: workspace files, session histories, evaluation questions, and staged update packages. Layer~5 is an internal generation guide for formatting and noise controls. Layer~0 is never shown, so verification is reliable while the observable layers behave like noisy reflections of the same reality.

\noindent \textbf{Multi-source conflict design.} ClawArena distinguishes integration (assembling a coherent picture from scattered evidence) from conflict reasoning (recognizing disagreement and deciding which sources dominate). Each scenario embeds four canonical evidence relations: factual conflicts (C1), authority conflicts (C2, undocumented or contradicted approvals), non-conflict slots (C3, genuine agreement), and temporal/process conflicts (C4). C3 blocks scoring well by treating every disagreement as a contradiction.

\noindent \textbf{Staged update design.} Updates arrive in stages: early rounds expose plausible but incomplete narratives, while later rounds inject contradictions and authoritative confirmations. Updates are subjective (appended messages shifting credibility) or objective (file modifications altering the factual record). Agents are scored on whether they revise once contradictions arrive, not on pre-emptive skepticism.

\noindent \textbf{Personalization protocol.} Personalization proceeds through four stages: (1)~calibration with natural hints (``put that in a table''); (2)~feedback via corrections; (3)~session-implicit, expressed only through interaction patterns; and (4)~silent-exam with no reminders; only stage-4 rounds are scored. Preferences span output format, artifact naming, document structure, analytical style, and communication tone.

\subsection{Construction pipeline}
\label{sec:construction}

A key challenge is producing scenarios at scale while preserving causal coherence across layers and realistic behavioral patterns. We use a three-stage pipeline combined with empirical distribution constraints (Figure~\ref{fig:pipeline}).

\noindent \textbf{Stage~1: Seed construction.} The first batch was authored by hand with cross-validation. For example, the startup outage scenario was iteratively refined until all four contradiction types were present, every answer required at least two sources, each update changed at least one previously correct answer, and all keys were unambiguous.

\noindent \textbf{Stage~2: Meta-specification induction.} From the seeds we distilled a meta-specification encoding structural invariants: narrative patterns, contradiction-type ratios, bias-phrase insertion rules, and update-question binding constraints (e.g., exactly one C3 slot per scenario to prevent over-flagging; at least one update must flip a previously asked answer). This parallels constraining large-scale generation with high-quality exemplars \citep{wang2022self,bai2022constitutional}.

\noindent \textbf{Stage~3: Batch generation with real-world grounding.} Over 200 published empirical distributions (email volume \citep{radicati2024email}, commit patterns \citep{github2024octoverse}, messaging activity \citep{golder2011diurnal}, social network structure \citep{dunbar1992neocortex,onnela2007structure}) constrain character profiles and generation along three authenticity axes. \emph{Workspace}: documents follow domain-specific conventions and resemble system exports. \emph{Session}: message timing follows scenario-specific diurnal curves, contact frequency follows a four-tier Dunbar-layer weighting \citep{dunbar1998social} where intimate contacts appear roughly $100\times$ more often than peripheral ones, and 30--50\% of messages are irrelevant noise. \emph{Causal}: all observable materials derive from a single Layer~0, preserving causal connections rather than independently fabricated text.

\noindent \textbf{Stage~4: Validation.} Every scenario is validated at three levels. \emph{Structural}: directory layout, question schema, file existence, session alternation, update integrity. \emph{Semantic}: contradiction coverage, answer-key consistency, and linkage between observable traces and Layer~0 provenance. \emph{Control}: bias-phrase placement and non-conflict-slot consistency. During development, these checks caught 37 specification errors before any evaluation began, so any observed failure reflects agent behavior rather than hidden scenario inconsistency.

\noindent \textbf{Stage~5: Refinement.} The released 12-scenario benchmark is a refined subset of an earlier larger candidate pool: scenarios that did not pass all three validation passes were removed, ambiguous answer keys were rewritten, contradiction-type coverage was tightened, and the multi-choice / executable-check ratio was rebalanced per scenario. The result is a smaller but structurally cleaner evaluation surface in which every retained scenario satisfies the full set of design constraints rather than approximating them at scale.

\subsection{Scoring}
\label{sec:scoring}

\noindent \textbf{Question-level scoring.}
\emph{Multi-choice} questions present 7--9 candidates; the agent selects a subset via a $\backslash$\texttt{bbox\{...\}} token. Credit is awarded only on exact-match; partial diagnostics (IoU, precision, recall, F1) are recorded but never contribute to the score, so selecting all options is not rewarded.
\emph{Executable checks} are binary pass/fail: a sandboxed shell command verifies whether a workspace claim holds in the actual files.

\noindent \textbf{Composite reliability metrics.} Alongside the Task Completion Rate (TCR, per-round mean score macro-averaged across scenarios, decomposed into MC and EC sub-scores), we report:

\emph{Success Cohesion (SC)} = $(S - k)/(N - 1)$, where $S$ correct rounds form $k$ consecutive success runs over $N$ rounds; SC = 1 for one unbroken streak, SC = 0 for pass/fail alternation even at 50\% TCR.

\emph{Failure Dispersion (FD)} = $1 - (S_f - k_f)/(N-1)$ for $S_f$ failures in $k_f$ runs; high FD indicates no prolonged collapse.

\emph{Robustness} $= \mathrm{SC} \times \mathrm{FD}$; the multiplicative form penalizes collapse on either axis.

\emph{Composite Reliability Score} $= (\mathrm{TCR} + \mathrm{Robustness})/2$. Derivations and worked examples are in Appendix~\ref{app:scoring}.

\section{Experiments}
\label{sec:experiments}

\subsection{Setup}

\noindent \textbf{Frameworks.} We evaluate five AI agent frameworks spanning a range of design complexity: OpenClaw (enterprise-grade, TypeScript), Claude Code (Anthropic's official CLI), NanoBot (minimalist, Python), PicoClaw (lightweight, Go), and MetaClaw (skill-driven self-evolving framework \citep{xia2026metaclaw}, built on top of OpenClaw as its executor). All are deployed as persistent assistants receiving unified conversational input on the same workspace. Architectural details are provided in Appendix~\ref{app:frameworks}.

\noindent \textbf{Models.} We evaluate 18 models spanning proprietary APIs, community-accessible providers, and self-hosted deployments. Proprietary models include GPT-5.5, GPT-5.4, GPT-5.2, GPT-5.1 (OpenAI), Kimi-K2.5 (Moonshot AI), and Claude Opus-4.7, Claude Sonnet-4.6, and Claude Haiku-4.5 (Anthropic). Community-accessible models via third-party APIs include GLM-5.1, GLM-4.7-Flash, Qwen3.6-Plus, DeepSeek-V4-Pro, Gemini-3.1-Pro, Mimo-V2.5-Pro, and Ling-2.6. Self-hosted models via local vLLM include Qwen3.6-35B, Qwen3.6-27B, and Gemma-4-31B. Cross-model comparisons fix the framework to OpenClaw except for the three Anthropic models, which are incompatible with OpenClaw and are therefore reported separately under Claude~Code; cross-framework comparisons are conducted under GPT-5.1.

\noindent \textbf{Evaluation protocol.} Each scenario consists of a sequence of rounds, and each round is a full multi-turn agent interaction whose API-call count varies from a single call to several dozen depending on tool use and reasoning depth, making evaluation far costlier than single-inference benchmarks. All reported experiments use the complete \textsc{ClawArena} benchmark (12 scenarios, 337 rounds, 45 dynamic updates).

\subsection{Cross-model comparison}
\label{sec:exp_model}

To isolate the effect of model capability, we fix the framework to OpenClaw and evaluate 15 models across the full benchmark (12 scenarios, 337 rounds), partitioned into proprietary API models and open-weight models. Table~\ref{tab:exp_model} shows results sorted by CRS within each group. We additionally report three Anthropic models (Claude Opus-4.7, Claude Sonnet-4.6, and Claude Haiku-4.5) in a separate \emph{provider-native} block because they are incompatible with OpenClaw and can only be exercised through Anthropic's Claude~Code harness; their numbers are therefore not directly comparable to the OpenClaw-hosted rows above and serve as reference points across the current Anthropic model tier.

\begin{table}[t]
\centering
\small
\caption{Cross-model comparison on 12 scenarios under OpenClaw. Models are partitioned into proprietary (top) and open-weight (middle), each sorted by CRS. The bottom block reports three Anthropic models under their provider-native Claude~Code harness (incompatible with OpenClaw); their numbers are not directly comparable with the OpenClaw rows above. \textbf{Bold} marks the best per column within each of the first two groups; \underline{underline} marks the runner-up. $^\dagger$~denotes self-hosted vLLM deployment; all others use provider APIs.}
\label{tab:exp_model}
% \resizebox{\linewidth}{!}{%
\begin{tabular}{@{}lcccccccc@{}}
\toprule
\small
\multirow{2}{*}{\textbf{Model}}
  & \multicolumn{3}{c}{\textbf{TCR}}
  & \multicolumn{3}{c}{\textbf{Robustness}}
  & \multirow{2}{*}{\textbf{CRS}} \\
\cmidrule(lr){2-4}\cmidrule(lr){5-7}
  & \textbf{Avg} & \textbf{MC} & \textbf{EC}
  & \textbf{Overall} & \textbf{SC} & \textbf{FD} & \\
\midrule
\multicolumn{8}{@{}l}{\textit{Proprietary}} \\[2pt]
GPT-5.5          & \textbf{78.34} & \textbf{75.79} & \textbf{79.34} & \textbf{58.22} & \textbf{61.24} & \underline{95.06} & \textbf{68.28} \\
GPT-5.1          & 70.33 & \textbf{75.79} & 68.18 & \underline{56.24} & \underline{58.96} & \textbf{95.37} & \underline{63.28} \\
GPT-5.4          & \underline{71.22} & \underline{71.58} & \underline{71.07} & 46.77 & 51.51 & 90.78 & 58.99 \\
Gemini-3.1-Pro   & 69.57 & 66.32 & \underline{71.07} & 45.62 & 50.54 & 90.23 & 57.59 \\
Qwen3.6-Plus     & 67.06 & \underline{71.58} & 65.29 & 43.29 & 47.89 & 90.38 & 55.17 \\
GPT-5.2          & 65.88 & 61.05 & 67.77 & 42.49 & 47.21 & 90.01 & 54.18 \\
\midrule
\multicolumn{8}{@{}l}{\textit{Open-weight}} \\[2pt]
Gemma-4-31B$^\dagger$   & \textbf{75.37} & \textbf{81.05} & \underline{73.14} & \textbf{52.22} & \textbf{56.76} & 91.90 & \textbf{63.80} \\
GLM-5.1                 & \underline{72.70} & \underline{72.63} & 72.73 & \underline{48.57} & \underline{52.74} & \underline{92.07} & \underline{60.63} \\
Mimo-V2.5-Pro           & 71.45 & 66.32 & \textbf{73.55} & 47.85 & 52.23 & 91.62 & 59.65 \\
Kimi-K2.5               & 69.44 & 60.00 & \underline{73.14} & 45.08 & 48.86 & 90.02 & 57.24 \\
Qwen3.6-27B$^\dagger$   & 66.63 & 65.26 & 68.60 & 45.07 & 48.40 & \textbf{93.12} & 55.85 \\
DeepSeek-V4-Pro         & 66.89 & 57.89 & 70.66 & 43.60 & 48.56 & 89.82 & 55.25 \\
Qwen3.6-35B-A3B$^\dagger$   & 60.24 & 51.58 & 63.64 & 37.49 & 42.17 & 88.93 & 48.86 \\
Ling-2.6                & 55.05 & 66.32 & 50.83 & 33.09 & 37.62 & 87.94 & 44.07 \\
GLM-4.7-Flash           & 54.10 & 42.11 & 57.02 & 23.54 & 30.55 & 77.05 & 38.82 \\
\midrule
\multicolumn{8}{@{}l}{\textit{Provider-native (Claude Code, not comparable to rows above)}} \\[2pt]
Claude Opus-4.7         & \textbf{76.13} & \textbf{65.26} & \textbf{80.58} & \textbf{57.47} & \textbf{60.06} & \textbf{94.06} & \textbf{66.31} \\
Claude Sonnet-4.6       & \underline{73.36} & 63.16 & \underline{77.69} & \underline{52.40} & \underline{54.80} & \underline{93.02} & \underline{62.16} \\
Claude Haiku-4.5        & 72.29 & \underline{64.21} & 75.62 & 50.61 & 54.74 & 90.54 & 60.93 \\
\bottomrule
\end{tabular}%
% }
\end{table}

Among proprietary models (all OpenClaw), GPT-5.5 leads on CRS (68.28) and TCR (78.34\%), with GPT-5.1 second (63.28). The proprietary range is 14.10 points, driven by the gap between the two GPT-5.x generations at the top and the remaining models clustered between 54 and 59 CRS. The provider-native block reports the Anthropic tier under Claude~Code: Opus-4.7 (CRS 66.31, TCR 76.13, EC 80.58), Sonnet-4.6 (CRS 62.16, TCR 73.36, EC 77.69), and Haiku-4.5 (CRS 60.93, TCR 72.29, EC 75.62). These rows are not directly comparable to the OpenClaw rows above since the harness differs, but the size-ordered hierarchy is preserved across all three metrics, and EC values consistently track the stronger proprietary tier, suggesting that workspace grounding remains the relative strength of the Claude family.

Among open-weight models, Gemma-4-31B leads on CRS (63.80) and dominates MC across the table (81.05\%), showing strong multi-choice reasoning is attainable at open-weight scale; GLM-5.1 ranks second (60.63). The open-weight range (24.98 points) is wider, reflecting greater architectural diversity, with the smaller Qwen3.6-27B posting the highest FD across the open-weight block (93.12) despite a mid-pack CRS, while GLM-4.7-Flash anchors the bottom (38.82 CRS). Gemma-4-31B comes within 4.5 CRS of GPT-5.5 but its EC trails by 6.2 points (73.14 vs.\ 79.34), indicating workspace grounding is the primary remaining gap.

MC and EC are only moderately correlated: DeepSeek-V4-Pro reaches 70.66 EC despite 57.89 MC, while Ling-2.6 shows the reverse (66.32 MC, 50.83 EC), so workspace grounding and reasoning are partially independent. The model-induced CRS range (29.46 points) exceeds the largest framework-induced range below (23.78 points under GPT-5.5), confirming that model capability dominates framework design.

\subsection{Cross-framework comparison}
\label{sec:exp_framework}

To isolate the effect of framework design, we evaluate four frameworks under three representative models (GPT-5.1, GPT-5.5, Kimi-K2.5) spanning two providers, examining whether framework-induced patterns generalize across model families.

\begin{table}[t]
\centering
\small
\caption{Cross-framework comparison. Three models are evaluated across all four frameworks. \textbf{Bold} marks the best per column within each model block; \underline{underline} marks the runner-up.}
\label{tab:exp_framework}
% \resizebox{\linewidth}{!}{%
\begin{tabular}{@{}llccccccc@{}}
\toprule
\multirow{2}{*}{\textbf{Model}} & \multirow{2}{*}{\textbf{Framework}}
  & \multicolumn{3}{c}{\textbf{TCR}}
  & \multicolumn{3}{c}{\textbf{Robustness}}
  & \multirow{2}{*}{\textbf{CRS}} \\
\cmidrule(lr){3-5}\cmidrule(lr){6-8}
  & & \textbf{Avg} & \textbf{MC} & \textbf{EC}
    & \textbf{Overall} & \textbf{SC} & \textbf{FD} & \\
\midrule
\multirow{4}{*}{GPT-5.5}
 & OpenClaw    & \textbf{78.34} & \textbf{75.79} & \textbf{79.34} & \textbf{58.22} & \textbf{61.24} & \textbf{95.06} & \textbf{68.28} \\
 & Claude Code & 59.31 & 67.37 & 55.79 & 30.30 & 36.27 & 81.84 & 44.50 \\
 & PicoClaw    & \underline{77.15} & \underline{74.74} & \underline{78.10} & \underline{56.66} & \underline{59.49} & \underline{94.74} & \underline{66.87} \\
 & NanoBot     & 73.00 & 72.63 & 73.14 & 50.98 & 55.24 & 90.76 & 61.88 \\
\midrule
\multirow{4}{*}{GPT-5.1}
 & OpenClaw    & \underline{70.33} & \textbf{75.79} & \underline{68.18} & \underline{56.24} & \underline{58.96} & \textbf{95.37} & \underline{63.28} \\
 & Claude Code & 55.19 & 65.26 & 51.24 & 39.95 & 43.66 & 91.54 & 47.57 \\
 & PicoClaw    & \textbf{71.51} & \underline{73.68} & \textbf{70.66} & \textbf{56.86} & \textbf{59.90} & \underline{94.90} & \textbf{64.18} \\
 & NanoBot     & 67.95 & \underline{73.68} & 65.70 & 51.80 & 55.09 & 94.04 & 59.88 \\
\midrule
\multirow{4}{*}{Kimi-K2.5}
 & OpenClaw    & \textbf{69.44} & \underline{60.00} & \textbf{73.14} & 45.08 & 48.86 & 90.02 & 57.24 \\
 & Claude Code & \underline{66.77} & 56.84 & \underline{70.66} & \textbf{52.73} & \textbf{56.26} & \textbf{93.74} & \textbf{59.75} \\
 & PicoClaw    & 65.58 & \textbf{65.26} & 65.70 & \underline{49.64} & \underline{53.72} & \underline{92.41} & \underline{57.61} \\
 & NanoBot     & 45.99 & 49.47 & 44.63 & 29.18 & 32.97 & 88.50 & 37.58 \\
\bottomrule
\end{tabular}%
% }
\end{table}

Under GPT-5.1, PicoClaw leads on CRS (64.18), with OpenClaw (63.28) close behind. OpenClaw retains the best MC (75.79), while PicoClaw leads on EC (70.66). The GPT-5.1 framework range is 16.61 points, narrower than the 29.46-point model range above.

The GPT-5.5 block shows a substantially larger framework swing across the same four frameworks: OpenClaw leads at 68.28 CRS with PicoClaw close behind (66.87), NanoBot trails at 61.88, and Claude Code collapses to 44.50, a 23.78-point intra-model spread. Both GPT-5.x configurations bottom out under Claude Code (47.57 for GPT-5.1, 44.50 for GPT-5.5), with the gap widening rather than narrowing as the model improves. This stands in sharp contrast to Kimi-K2.5, where Claude Code is in fact the strongest framework (59.75 CRS), suggesting that the OpenAI GPT family interacts poorly with the Anthropic Claude Code harness, whose tool-binding and prompt scaffold appear tuned to Claude-style instruction following.

The Kimi-K2.5 block shows framework effects generalize across families but differ in magnitude: Claude Code achieves the best CRS for Kimi-K2.5 (59.75), suggesting its workspace-native design compensates for structured-response instability elsewhere, while NanoBot lags (37.58), a drop more than twice the analogous NanoBot/PicoClaw gap under GPT-5.1. Framework robustness interacts non-linearly with model reliability.

\subsection{Skill-driven self-evolution}
\label{sec:exp_metaclaw}

To isolate skill-based self-evolution, we fix model and framework and compare each model against its MetaClaw-enabled variant (Table~\ref{tab:exp_metaclaw}, four matched pairs).

\begin{table}[t]
\centering
\small
\caption{Skill-driven self-evolution ablation. Each pair shares the same model and base OpenClaw configuration; the +MetaClaw variant adds skill injection. $\Delta$CRS is the change relative to the baseline.}
\label{tab:exp_metaclaw}
% \resizebox{\linewidth}{!}{%
\begin{tabular}{@{}llccccccc@{}}
\toprule
\multirow{2}{*}{\textbf{Model}} & \multirow{2}{*}{\textbf{Config}}
  & \multicolumn{3}{c}{\textbf{TCR}}
  & \multicolumn{2}{c}{\textbf{Robustness}}
  & \multirow{2}{*}{\textbf{CRS}} \\
\cmidrule(lr){3-5}\cmidrule(lr){6-7}
  & & \textbf{Avg} & \textbf{MC} & \textbf{EC} & \textbf{SC} & \textbf{FD} & \\
\midrule
\multirow{2}{*}{GPT-5.5}
 & Baseline         & 78.34 & 75.79 & 79.34 & 61.24 & 95.06 & 68.28 \\
 & +MetaClaw skills & 78.86 & 75.79 & 80.58 & 61.90 & 95.19 & 68.89 \textcolor{gray}{(+0.61)} \\
\midrule
\multirow{2}{*}{GPT-5.1}
 & Baseline         & 70.33 & 75.79 & 68.18 & 58.96 & 95.37 & 63.28 \\
 & +MetaClaw skills & 70.62 & 74.74 & 69.01 & 59.05 & 96.22 & 63.72 \textcolor{gray}{(+0.44)} \\
\midrule
\multirow{2}{*}{GLM-5.1}
 & Baseline         & 72.70 & 72.63 & 72.73 & 52.74 & 92.07 & 60.63 \\
 & +MetaClaw skills & 73.29 & 69.47 & 74.79 & 54.65 & 92.19 & 61.82 \textcolor{gray}{(+1.19)} \\
\midrule
\multirow{2}{*}{Qwen3.6-Plus}
 & Baseline         & 67.06 & 71.58 & 65.29 & 47.89 & 90.38 & 55.17 \\
 & +MetaClaw skills & 67.06 & 68.42 & 66.53 & 48.41 & 90.72 & 55.50 \textcolor{gray}{(+0.33)} \\
\bottomrule
\end{tabular}%
% }
\end{table}

Across the four pairs, MetaClaw improves CRS by 0.33--1.19 without degrading TCR. The primary mechanism is Robustness: SC and FD both rise in every pair, so skill injection reduces isolated failures and suppresses long failure streaks rather than changing the raw correct-answer rate. The largest gain occurs for GLM-5.1 (+1.19 CRS, +1.79 Robustness) whose baseline Robustness is least consistent; the smallest (Qwen3.6-Plus, +0.33 CRS) on the most stable baseline. The strongest baseline (GPT-5.5, 68.28 CRS) still benefits (+0.61), confirming the overlay is not redundant for already-capable models. MetaClaw is a reliable but modest enhancer that narrows the gap to stronger baselines through consistency, not accuracy.

\subsection{Error analysis}
\label{sec:exp_error}

We characterize agent failures with ten per-option case studies covering the MS, DU, P, and exec\_check dimensions; scenario-specific evidence and round-level diagnostics are deferred to Appendix~\ref{app:casestudies} (Figures~\ref{fig:casestudy_01_02}--\ref{fig:casestudy_09_10}). Six recurring failure modes emerge.

\noindent \textbf{Compound claims are omitted while primary facts are captured.} MS-R errors concentrate on options that require conjoining two clauses of the same source or promoting a cross-statement relation to a first-class claim. When such compound options appear, exact-match rates collapse even though every atomic fact is correctly recalled; the ceiling is therefore set by willingness to aggregate, not by evidence availability.

\noindent \textbf{Belief revision fails asymmetrically across update packages.} DU-R failures cluster on specific update installments rather than accumulating monotonically with update count. The decisive factor is the bridging inference each update demands (e.g., distinguishing partial from full restitution); updates that merely extend prior evidence are handled reliably while those that force a re-interpretation of earlier claims are not.

\noindent \textbf{Framework choice does not compensate for model-level output-format gaps.} When the same model fails a strict-schema exec\_check task under one framework, it fails under the alternate framework with identical schema-compliance errors, even as other models clear the same task under the first framework. Framework contribution is substantive for in-context reasoning discipline but limited for literal structural output.

\noindent \textbf{Compound structural constraints define a capability ceiling.} A small cluster of exec\_check rounds attains 0/8 pass rates across all eight configurations. These rounds share a common profile: correct factual content plus a conjunction of structural constraints (timezone-aware timestamps, typed JSON fields with provenance arrays, enum-restricted labels, exact top-level key names). The underlying facts are usually identified; the failure is compound output-format compliance that MC metrics cannot detect.

\noindent \textbf{Implicit-preference compliance decays as checkers tighten.} Models that pass a lenient mid-scenario preference check often fall off once a later round layers additional literal constraints onto the same convention. Passing a softer preference check does not predict compliance with a stricter successor, indicating brittle rather than internalized preference tracking.

\noindent \textbf{MC and EC skills remain loosely correlated.} Set-selection leaders do not dominate exec\_check, and models with sporadic MC extraction failures can still succeed on script-generation rounds when the expected schema is listed verbatim. Workspace grounding draws on file parsing, shell construction, and literal-key discipline, which are only weakly predicted by multi-choice reasoning accuracy, supporting the two-format design of \textsc{ClawArena}. Performance variation across domains exceeds 60\% for every model tested; the Chinese-language enterprise scenario disproportionately favors models with stronger multilingual training, and the hardest scenario keeps every configuration below 30\% CRS.

\section{Conclusion}

We introduced \textsc{ClawArena}, a benchmark for AI agents in evolving information environments. Across five frameworks and 18 models, model capability dominates framework design in CRS range (29 vs.\ 24 points), MetaClaw's skill overlay improves Robustness without degrading accuracy across all four tested families, belief revision difficulty is governed by update specificity rather than volume, and per-option diagnostics expose failure modes that aggregate scores conceal. A natural extension moves beyond staged updates toward live, unconstrained environments where agents formulate their own queries against real-time sources.

\bibliographystyle{colm2026_conference}
\bibliography{colm2026_conference}

\newpage
\appendix

\onecolumn
\startcontents[appendix]
\printcontents[appendix]{ }{0}{\section*{Appendix}}

\section{Data Sample Overview}
\label{app:domain_gallery}

\begin{figure}[ht]
\centering
\includegraphics[width=\textwidth]{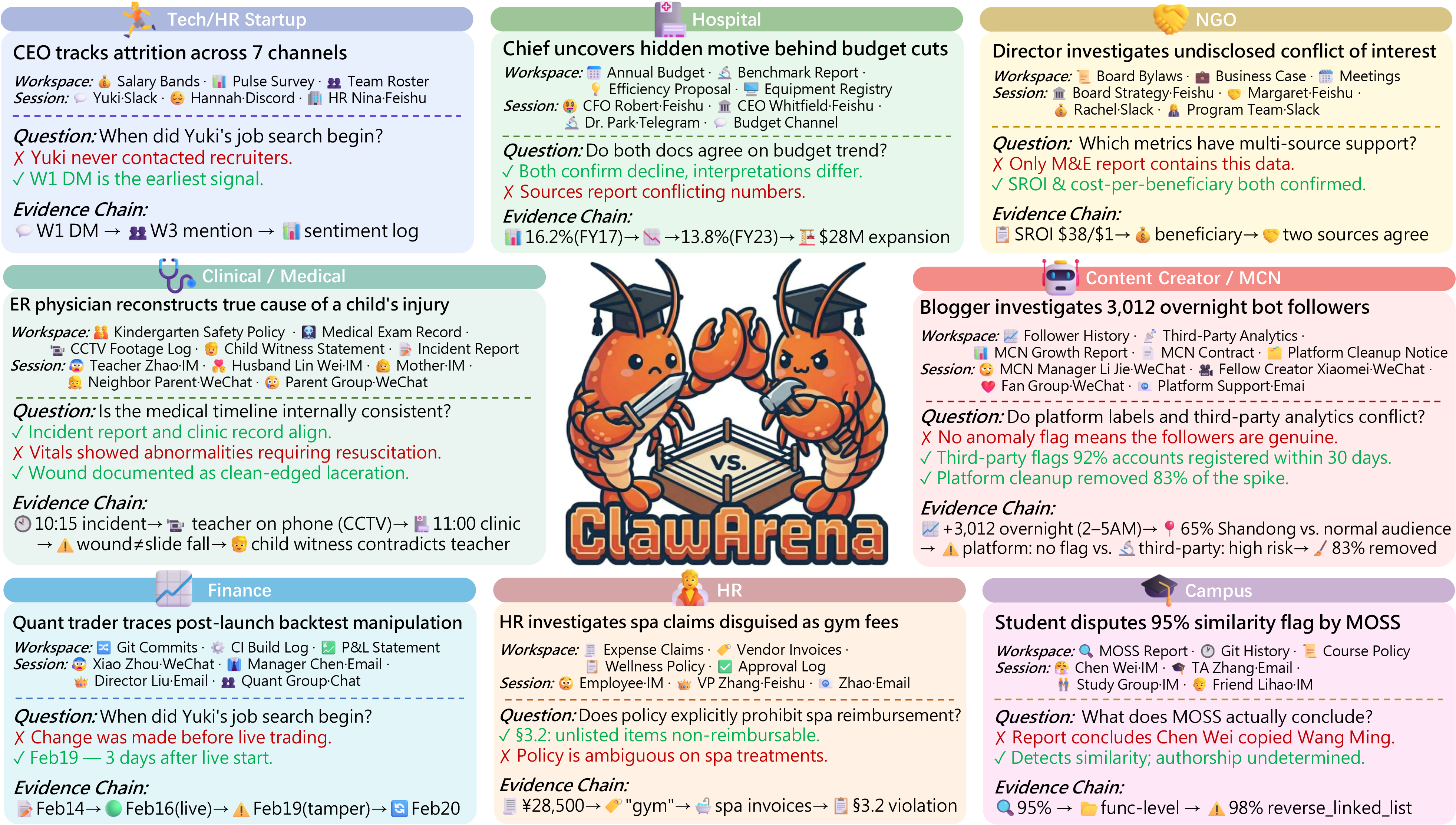}
\caption{Cross-domain data sample gallery from \textsc{ClawArena}. Each tile presents one scenario in a distinct professional context, highlighting the workspace, session sources, evaluation question, and an evidence chain that traverses conflicting and time-evolving observations. The center logo reflects the benchmark's adversarial spirit: agents must ``claw'' through conflicting evidence to reach the ground truth.}
\label{fig:domain_gallery}
\end{figure}

Figure~\ref{fig:domain_gallery} illustrates the surface form of \textsc{ClawArena} scenarios across the diverse professional contexts represented in the benchmark. Each tile shows the workspace, multi-channel session sources, the type of evaluation question posed, and a representative evidence chain through which the agent must navigate to recover the hidden ground truth. The gallery is intended as a qualitative orientation to the data style of \textsc{ClawArena}; the structural framework, evaluation dimensions, and scoring summarized in Figure~\ref{fig:overview} are formalized in Section~\ref{sec:clawarena}.

\section{Framework Details}
\label{app:frameworks}

We evaluate five AI agent frameworks with representative design differences, all deployed as persistent assistants receiving unified conversational input and operating on the same workspace environment.

\noindent \textbf{OpenClaw} (TypeScript, Agent Client Protocol architecture) provides enterprise-grade session management and tool binding. It supports multi-channel session routing, structured memory with semantic retrieval, and configurable tool pipelines.

\noindent \textbf{MetaClaw} (skill-driven self-evolving framework) maintains a growing repository of procedural skills distilled from prior failure trajectories; these skills are retrieved and injected into the agent's prompt on each round without modifying model weights. In this work it is built on top of OpenClaw as its executor, so that differences in downstream behaviour isolate the effect of skill injection over the same tool-binding and memory stack.

\noindent \textbf{Claude Code} (Anthropic's official CLI) is optimized for code and file workspace operations with terminal-native interaction. It uses a single long-context window and relies on the model's in-context reasoning rather than external memory modules.

\noindent \textbf{NanoBot} (Python, $\sim$1,700 lines) follows a minimalist design with a \texttt{MEMORY.md} + \texttt{HISTORY.md} dual-layer memory. Its simplicity makes it a useful lower-bound baseline for framework complexity.

\noindent \textbf{PicoClaw} (Go, $<$10\,MB runtime memory) uses an event-driven architecture with date-organized memory. It prioritizes low resource consumption over retrieval sophistication.

\section{Implementation Details}
\label{app:implementation}

\noindent \textbf{Model providers.} Table~\ref{tab:impl_providers} lists the provider routing for every model evaluated in Section~\ref{sec:experiments}. Proprietary and community-accessible models are served through their respective commercial APIs; self-hosted models are deployed locally via vLLM. Claude~Haiku-4.5 is the sole model that cannot be served into OpenClaw's tool-binding stack, and is instead exercised through Anthropic's provider-native Claude~Code harness using a Claude OAuth credential. No third-party API endpoint is used for this model.

\begin{table}[ht]
\centering
\small
\caption{Provider routing for all evaluated models.}
\label{tab:impl_providers}
\begin{tabular*}{\linewidth}{@{\extracolsep{\fill}}lll@{}}
\toprule
\textbf{Provider} & \textbf{Access} & \textbf{Models} \\
\midrule
OpenAI            & Codex API           & GPT-5.5, GPT-5.4, GPT-5.2 \\
Azure OpenAI      & Azure endpoint      & GPT-5.1, Kimi-K2.5 \\
Anthropic         & Claude OAuth        & Claude Opus-4.7, Claude Sonnet-4.6, Claude Haiku-4.5 \\
OpenRouter        & OpenRouter API      & GLM-5.1 (\texttt{z-ai}), GLM-4.7-Flash (\texttt{z-ai}), Qwen3.6-Plus \\
                  &                     & (\texttt{qwen}), DeepSeek-V4-Pro (\texttt{deepseek}), Gemini-3.1-Pro \\
                  &                     & (\texttt{google}), Mimo-V2.5-Pro (\texttt{xiaomi}), Ling-2.6 \\
                  &                     & (\texttt{inclusionai}) \\
Self-hosted vLLM  & Local (4$\times$GPU) & Qwen3.6-35B-A3B, Qwen3.6-27B, Gemma-4-31B-it \\
\bottomrule
\end{tabular*}
\end{table}

\noindent \textbf{Self-hosted vLLM deployment.} The two open-weight models evaluated locally are served with vLLM on 4$\times$NVIDIA RTX 6000 Ada Generation GPUs (48\,GB each) under tensor parallelism of 4. Both share a maximum sequence length of 204{,}800 tokens and a GPU memory utilization target of 0.92. Model-specific parser and quantization settings are summarized in Table~\ref{tab:impl_vllm}.

\begin{table}[ht]
\centering
\small
\caption{vLLM serving configuration for self-hosted models. TP denotes tensor parallelism; \texttt{max-model-len} is the served context length; \texttt{tool-call-parser} and \texttt{reasoning-parser} select the vLLM output-format adapters.}
\label{tab:impl_vllm}
\begin{tabular*}{\linewidth}{@{\extracolsep{\fill}}lcccccl@{}}
\toprule
\textbf{Model} & \textbf{TP} & \textbf{max-model-len} & \textbf{tool-call-parser} & \textbf{reasoning-parser} & \textbf{KV cache} \\
\midrule
Qwen3.6-35B-A3B & 4 & 204{,}800 & \texttt{qwen3\_coder} & \texttt{qwen3}  & bf16 \\
Gemma-4-31B-it  & 4 & 204{,}800 & \texttt{gemma4}       & \texttt{gemma4} & fp8  \\
\bottomrule
\end{tabular*}
\end{table}

\section{Scoring Details}
\label{app:scoring}

\noindent \textbf{Multi-choice (set-selection).} Each question presents 7--9 candidate statements; the agent must select the correct subset by enclosing its answer in a $\backslash$\texttt{bbox\{\}} token. Credit is awarded only when the selected set matches the gold set exactly (exact-match). Partial diagnostics (IoU, precision, recall, and F1) are recorded per round for analysis but do not contribute to any reported score, ensuring that selecting all options to guarantee a non-zero partial credit is never rewarded.

\noindent \textbf{Executable checks.} Executable-check questions are binary pass/fail: a sandboxed shell command tests whether a specific workspace-level claim (e.g., that a file contains a particular entry, or that two documents report consistent timestamps) holds in the actual files. Executable-check scores are independent of multi-choice scores and test workspace grounding rather than reasoning.

\noindent \textbf{Composite Reliability Score (CRS).}
Let a configuration produce a sequence of $N$ round scores $s_1, \dots, s_N \in \{0, 1\}$. Define $S = \sum_i s_i$ (correct rounds), $k$ = number of maximal consecutive success runs, $S_f = N - S$ (failed rounds), and $k_f$ = number of maximal consecutive failure runs. The four aggregate metrics are:

\vspace{0.5em}
\noindent\textbf{TCR} $= S / N$.\quad Mean correctness across all rounds.

\noindent\textbf{SC (Success Cohesion)} $= (S - k)/(N - 1)$ if $N > 1$, else 0.\quad Measures how much correct performance is concentrated into long unbroken runs. A configuration that passes all rounds in one streak achieves SC~$= 1$; one that alternates pass/fail achieves SC~$= 0$ even at 50\% TCR.

\noindent\textbf{FD (Failure Dispersion)} $= 1 - (S_f - k_f)/(N-1)$ if $N > 1$, else 1.\quad Measures how evenly failures are spread across the sequence. High FD indicates no prolonged collapse; a single long failure run drives FD toward 0.

\noindent\textbf{Robustness} $= \mathrm{SC} \times \mathrm{FD}$.\quad Multiplicative form ensures that a collapse on either axis (many isolated successes or one long failure streak) substantially reduces the score.

\noindent\textbf{CRS} $= (\mathrm{TCR} + \mathrm{Robustness}) / 2$.\quad Gives equal weight to raw correctness and behavioral consistency. All metrics are macro-averaged across scenarios before CRS is computed.

\vspace{0.5em}
\noindent\textbf{Worked example.} For $N = 5$ with scores $[1,0,1,0,1]$: $S = 3$, $k = 3$, $S_f = 2$, $k_f = 2$. Then SC $= (3-3)/4 = 0$, FD $= 1 - (2-2)/4 = 1$, Robustness $= 0$, TCR $= 0.6$, CRS $= 0.3$. Although TCR is 60\%, the alternating pattern yields zero Robustness, pulling CRS well below TCR. For scores $[1,1,1,0,0]$: $S = 3$, $k = 1$, $S_f = 2$, $k_f = 1$. SC $= (3-1)/4 = 0.5$, FD $= 1 - (2-1)/4 = 0.75$, Robustness $= 0.375$, CRS $= 0.488$. The same TCR but coherent success and contained failure yields a substantially higher CRS.

\section{Benchmark Statistics}
\label{app:core_stats}

This section characterizes \textsc{ClawArena} in full statistical detail, covering token composition, question shape, staged-update structure, and per-scenario breakdowns referenced throughout Section~\ref{sec:experiments}. Cross-model comparisons fix the framework to OpenClaw; cross-framework comparisons fix the model to GPT-5.1. Each round is a full multi-turn agent interaction whose API-call count varies from a single call to several dozen depending on tool use and reasoning depth.

\noindent \textbf{Tokenization.} Token counts are produced with OpenAI's \texttt{cl100k\_base} byte-pair-encoding tokenizer (the encoding used by GPT-4/GPT-4o and by \texttt{tiktoken} as a cross-model reference). For plain artefacts (workspace files, question strings, feedback strings, markdown transcripts) the raw UTF-8 bytes are encoded directly. For jsonl session files (whose on-disk layout differs sharply across frameworks) we first parse the transcript into a normalized sequence of \texttt{\{role, content\}} messages, concatenating text parts and flattening tool-call arguments, and then re-render the sequence into the ChatML template (\texttt{<|im\_start|>role\textbackslash n content<|im\_end|>}) before tokenization. This parse-then-render step approximates the prompt payload that the inference backend assembles from the transcript, rather than charging the model for the per-record JSON braces, keys (\texttt{role}, \texttt{content}, \texttt{timestamp}, \ldots) and metadata that each framework happens to persist on disk. Markdown transcripts (used by Claude Code and NanoBot for per-turn history) are already in near-prompt form and are counted verbatim. The same tokenizer is applied uniformly across all frameworks and artefact types, so the per-framework differences reported in Table~\ref{tab:core_tokens} reflect genuine differences in session-format verbosity (tool-call encoding, per-turn metadata volume, history containerization) rather than measurement artefacts.

\subsection{Overall scale}

\textsc{ClawArena} spans 12 scenarios and 337 rounds. Of these, 45 rounds (13.4\%) expose at least one staged update and 24 rounds (7.1\%) carry an implicit user preference. Per-scenario round counts range from 24 to 30 (mean 28.1), giving balanced coverage across the represented professional contexts. Under the OpenClaw layout, the cumulative input across the benchmark totals 4{,}739{,}550 tokens, dominated by workspace content; the other three frameworks fall within a 0.4\% band of this figure (see Table~\ref{tab:core_tokens}).

\subsection{Token composition}

Table~\ref{tab:core_tokens} decomposes the total token budget by source under each of the four compatible frameworks. Workspace content (baseline files plus staged workspace updates) accounts for 90.1--90.3\% of tokens across frameworks, reflecting the benchmark's focus on file-grounded reasoning rather than long dialogue histories. Session-side updates contribute a further 2.6\% of tokens, with the three jsonl-based frameworks (Claude Code, NanoBot, PicoClaw) carrying marginally larger session-update budgets than OpenClaw because their tool-call argument payloads survive the parse-and-render step intact, whereas Claude Code and NanoBot transcribe per-turn history as markdown and PicoClaw as jsonl into the workspace message-log tree. The entire authored conversational surface (questions, feedback, explicit preferences) consumes only 3.5\% of tokens yet drives every scored interaction.

\begin{table}[ht]
\centering
\caption{Token distribution across input sources in \textsc{ClawArena}, reported independently for each of the four compatible frameworks. All counts use the \texttt{cl100k\_base} tokenizer; jsonl session files are parsed into a normalized message sequence and rendered into the ChatML template before counting, while markdown transcripts and all non-session artefacts are counted verbatim (see Appendix~\ref{app:core_stats} narrative).}
\label{tab:core_tokens}
\resizebox{\linewidth}{!}{%
\begin{tabular}{@{}l rr rr rr rr@{}}
\toprule
\multirow{2}{*}{\textbf{Source}}
  & \multicolumn{2}{c}{\textbf{OpenClaw}}
  & \multicolumn{2}{c}{\textbf{Claude Code}}
  & \multicolumn{2}{c}{\textbf{NanoBot}}
  & \multicolumn{2}{c}{\textbf{PicoClaw}} \\
\cmidrule(lr){2-3}\cmidrule(lr){4-5}\cmidrule(lr){6-7}\cmidrule(lr){8-9}
& \textbf{Tokens} & \textbf{\%}
& \textbf{Tokens} & \textbf{\%}
& \textbf{Tokens} & \textbf{\%}
& \textbf{Tokens} & \textbf{\%} \\
\midrule
Main session         & 10{,}457     & 0.221 & 4{,}170      & 0.088 & 7{,}976      & 0.168 & 7{,}989      & 0.168 \\
History sessions     & 158{,}090    & 3.335 & 167{,}139    & 3.531 & 167{,}139    & 3.519 & 159{,}359    & 3.361 \\
Workspace (base)     & 1{,}483{,}834 & 31.307 & 1{,}472{,}224 & 31.103 & 1{,}484{,}686 & 31.260 & 1{,}484{,}566 & 31.305 \\
Workspace update     & 2{,}797{,}037 & 59.015 & 2{,}797{,}133 & 59.095 & 2{,}797{,}037 & 58.891 & 2{,}797{,}037 & 58.983 \\
Session update       & 122{,}271    & 2.580 & 124{,}800    & 2.637 & 124{,}800    & 2.628 & 125{,}343    & 2.643 \\
Questions            & 101{,}881    & 2.149 & 101{,}881    & 2.152 & 101{,}881    & 2.145 & 101{,}881    & 2.148 \\
Feedback             & 64{,}271     & 1.356 & 64{,}271     & 1.358 & 64{,}271     & 1.353 & 64{,}271     & 1.355 \\
Preferences          & 1{,}709      & 0.036 & 1{,}709      & 0.036 & 1{,}709      & 0.036 & 1{,}709      & 0.036 \\
\midrule
Total                & 4{,}739{,}550 & 100.000 & 4{,}733{,}327 & 100.000 & 4{,}749{,}499 & 100.000 & 4{,}742{,}155 & 100.000 \\
\bottomrule
\end{tabular}%
}
\end{table}

\subsection{Question statistics}

The 337 rounds partition into 242 executable-check (exec\_check, 71.8\%) and 95 multi-choice (multi\_choice, 28.2\%) rounds. All multi-choice rounds are multi-answer. Table~\ref{tab:core_mc_shape} summarizes the shape of multi-choice prompts and the feature coverage of executable-check prompts.

\begin{table}[ht]
\centering
\small
\caption{Multi-choice prompt shape and executable-check feature coverage in \textsc{ClawArena}. MC statistics are computed over all 95 multi-choice rounds; EC statistics are computed over all 242 executable-check rounds.}
\label{tab:core_mc_shape}
\begin{tabular*}{\linewidth}{@{\extracolsep{\fill}}llccc@{}}
\toprule
\textbf{Dimension} & \textbf{Metric} & \textbf{Mean} & \textbf{Min} & \textbf{Max} \\
\midrule
Multi-choice      & Options per round                    & 6.38  & 5   & 10 \\
                  & Correct answers per round            & 4.46  & 2   & 8  \\
\midrule
Executable-check  & Sandbox timeout (s)                  & 40.2  & 30  & 60 \\
                  & \texttt{expect\_exit} coverage (\%)  & 100.0 & --- & --- \\
                  & \texttt{expect\_stdout} coverage (\%)& 0.4   & --- & --- \\
\bottomrule
\end{tabular*}
\end{table}

Every multi-choice round requires selecting a subset rather than a single option, eliminating single-answer degeneracy; the mean 4.46-of-6.38 selection ratio makes random guessing expensively uninformative. Executable-check rounds universally assert a specific exit status against a bounded timeout, with \texttt{expect\_stdout} used sparingly (1/242) so that pass/fail outcomes hinge on workspace state rather than output string formatting. Implicit preferences appear in 24 rounds (7.1\%), concentrated in P-R-category scenarios.

\subsection{Staged updates}

Across the 45 update-bearing rounds there are 83 update events touching 25 distinct workspace files. Table~\ref{tab:core_updates} breaks down the update events by target scope and the file-level actions by type.

\begin{table}[ht]
\centering
\small
\caption{Staged-update distribution. \emph{Target} reports the 83 update events; \emph{Action} reports the 25 file-level update operations triggered by those events.}
\label{tab:core_updates}
\begin{tabular*}{\linewidth}{@{\extracolsep{\fill}}llrr@{}}
\toprule
\textbf{Facet} & \textbf{Category} & \textbf{Count} & \textbf{\%} \\
\midrule
Target (events) & Workspace-targeted & 42 & 50.6 \\
                & Session-targeted   & 37 & 44.6 \\
                & Group-targeted     & 4  & 4.8  \\
\midrule
Action (files)  & \texttt{append}    & 14 & 56.0 \\
                & \texttt{new}       & 11 & 44.0 \\
\bottomrule
\end{tabular*}
\end{table}

The roughly balanced split between workspace- and session-targeted updates exercises both belief-revision pathways tested by DU-R: modifying or extending evidence in the filesystem versus introducing new turns into the conversation history. File-level actions are similarly balanced between \texttt{append} (extending existing artefacts) and \texttt{new} (introducing previously unseen files), ensuring that neither pure accumulation nor pure discovery dominates the revision workload. Per-round file mutations average 0.30, with a maximum of four files modified in a single round.

\subsection{Per-scenario breakdown}

Table~\ref{tab:core_per_scenario} lists round counts, question-type mix, preference and update coverage, update and workspace file counts, and total tokens for each of the 12 scenarios under the OpenClaw layout. All scenarios fall within a narrow 24--30 round band, but token totals vary by nearly two orders of magnitude due to one outlier scenario (\texttt{hil\_s1}) whose staged workspace updates inject the bulk of the benchmark's workspace tokens.

\begin{table}[ht]
\centering
\small
\caption{Per-scenario breakdown under the OpenClaw layout. Columns: rounds, multi-choice rounds (MC), executable-check rounds (EC), rounds with a preference (w/Pref), rounds exposing updates (w/Upd), total update events (Upd), files touched by updates (UpdF), baseline workspace files (WSF, excluding framework-injected session transcripts), and total tokens.}
\label{tab:core_per_scenario}
\resizebox{\linewidth}{!}{%
\begin{tabular}{@{}lrrrrrrrrr@{}}
\toprule
\textbf{Scenario} & \textbf{Rounds} & \textbf{MC} & \textbf{EC} & \textbf{w/Pref} & \textbf{w/Upd} & \textbf{Upd} & \textbf{UpdF} & \textbf{WSF} & \textbf{Tokens} \\
\midrule
\texttt{hil\_s1} & 24 & 8 & 16 & 3 & 4 & 8  & 25 & 24 & 4{,}344{,}165 \\
\texttt{hil\_c7} & 28 & 8 & 20 & 2 & 3 & 6  & 9  & 12 & 44{,}210 \\
\texttt{hil\_d3} & 30 & 8 & 22 & 2 & 4 & 8  & 10 & 12 & 44{,}137 \\
\texttt{hil\_e4} & 24 & 7 & 17 & 2 & 3 & 6  & 6  & 11 & 40{,}112 \\
\texttt{hil\_g4} & 27 & 8 & 19 & 2 & 3 & 7  & 8  & 10 & 37{,}429 \\
\texttt{hil\_g1} & 30 & 8 & 22 & 3 & 4 & 7  & 8  & 10 & 36{,}077 \\
\texttt{hil\_h3} & 27 & 8 & 19 & 2 & 4 & 6  & 7  & 10 & 35{,}038 \\
\texttt{hil\_j1} & 30 & 8 & 22 & 2 & 4 & 5  & 6  &  8 & 34{,}823 \\
\texttt{hil\_f3} & 30 & 8 & 22 & 2 & 4 & 6  & 7  & 11 & 34{,}061 \\
\texttt{hil\_i2} & 30 & 8 & 22 & 2 & 4 & 7  & 7  & 10 & 30{,}940 \\
\texttt{hil\_g3} & 30 & 8 & 22 & 0 & 4 & 6  & 7  &  7 & 30{,}474 \\
\texttt{hil\_f7} & 27 & 8 & 19 & 2 & 4 & 11 & 7  & 10 & 28{,}084 \\
\bottomrule
\end{tabular}%
}
\end{table}

\texttt{hil\_s1} alone accounts for 91.7\% of benchmark tokens, driven chiefly by its staged workspace updates (25 files) atop an already-large 24-file baseline; the remaining 11 scenarios collectively consume 395{,}385 tokens (8.3\%) and share a compact 7--12-file workspace baseline. This bimodal token distribution is intentional: it preserves a representative workspace-heavy stress scenario while keeping the aggregate evaluation cost tractable. Baseline workspace file counts (WSF) are computed from the OpenClaw layout after excluding framework-injected session-equivalent directories; the three non-native-multi-session frameworks (Claude Code, NanoBot, PicoClaw) instead transcribe per-turn history into \texttt{message\_logs/} and \texttt{sessions/} subtrees, which are classified as session content rather than workspace baseline and so do not inflate WSF. \texttt{hil\_f7} carries the densest update schedule (11 update events in 27 rounds) with 7 update-touched files atop a 10-file baseline, isolating the effect of high-frequency session-side belief revision. The multi-choice and executable-check quotas per scenario are nearly uniform (MC $\in \{7,8\}$, EC $\in \{16,22\}$), preventing any single scenario from biasing the exec\_check-dominated question distribution reported in the previous subsection.

\section{Meta-Specification Templates}
\label{app:metaspec}

Each ClawArena scenario is produced from a meta-specification that encodes structural invariants across seven template documents: the execution guide, narrative bible (Layer~0), evidence emission map, and Layers~1--4. Below we present all seven templates (Section~\ref{app:templates}).

\definecolor{boxblue}{RGB}{44,82,130}
\definecolor{boxbluebg}{RGB}{237,242,248}

\subsection{Template 1: Execution Guide}
\label{app:templates}

The execution guide is the entry point for scenario construction, linking all six layers and defining the build workflow.

\begin{tcolorbox}[breakable,colback=boxbluebg,colframe=boxblue,title={\footnotesize\bf Execution Guide Template (GUIDE.md)},fonttitle=\sffamily,coltitle=white,colbacktitle=boxblue,boxrule=0.5pt,arc=2pt]
\footnotesize
\textbf{1.\ Task Overview}\quad Task ID / Domain / Core evaluation goals / Final output target

\textbf{2.\ Spec File Index}\\
\begin{tabular}{@{\quad}lll}
\texttt{layer0-narrative.md} & truth baseline & read first \\
\texttt{layer1-workspace.md} & workspace plan & read second \\
\texttt{layer2-sessions.md} & session plan & read third \\
\texttt{layer3-eval.md} & round plan & read fourth \\
\texttt{layer4-dynamic.md} & update plan & read fifth \\
\end{tabular}

\textbf{3.\ Role and Session Table}\quad Each row maps a character role to a communication channel, session filename, and whether it appears in the initial release or a later update.

\textbf{4.\ Contradiction and Bias Quick-Reference}\quad Lists every contradiction (C1--C4) and bias (B1--B2) with the round where it first becomes visible and the round where reversal evidence arrives.

\textbf{5.\ Eight-Step Execution Workflow}\\
\quad Step~0: Read all layers, generate UUIDs, freeze filenames.\quad Step~1: Create fixed agent files + initial workspace files.\\
\quad Step~2: Write session intermediate JSON files.\quad Step~3: Build \texttt{.jsonl} session files from intermediate JSON.\\
\quad Step~4: Create update source files under \texttt{updates/}.\quad Step~5: Write \texttt{questions.json}.\\
\quad Step~6: Register sessions and update metadata.\quad Step~7: Append to \texttt{all\_tests.json}.\\
\quad Step~8: Run structural, semantic, and control validation.

\textbf{6.\ Mandatory Checks}\quad Data text is English; \texttt{questions.json} is a single group object; update-created sessions carry a \texttt{channel} field; initial sessions are registered in \texttt{sessions.json}.
\end{tcolorbox}
\vspace{-1mm}
\noindent{\footnotesize\textit{Table~\ref*{tab:guide_template}: Execution guide template, linking all six specification layers and defining the build workflow.}\refstepcounter{table}\label{tab:guide_template}}

\subsection{Template 2: Narrative Bible (Layer 0)}

Layer~0 defines the hidden ground truth that is never shown to the evaluated system. It structures the objective timeline, role-level truth gaps, contradiction map, bias design, and evaluation traps.

\begin{tcolorbox}[breakable,colback=boxbluebg,colframe=boxblue,title={\footnotesize\bf Narrative Bible Template (Layer 0)},fonttitle=\sffamily,coltitle=white,colbacktitle=boxblue,boxrule=0.5pt,arc=2pt]
\footnotesize
\textbf{1.\ Scene Summary}\quad Task ID / Domain / Time span / Main protagonist / Core benchmark factors (MS, DU, P).

\textbf{2.\ Objective Timeline}\quad Each row: \textit{Time} $\mid$ \textit{Objective event} $\mid$ \textit{What actually happened} $\mid$ \textit{Who knew at that time}.

\textbf{3.\ Role-Level Truth vs Self-Narrative}\quad For each character: \textit{Objective position}, \textit{Public narrative}, \textit{Private narrative}, and \textit{Why the gap exists}.

\textbf{4.\ Contradiction Map}\quad Each row: \textit{ID} $\mid$ \textit{Description} $\mid$ \textit{Source~A claim + location} $\mid$ \textit{Source~B claim + location} $\mid$ \textit{Objective truth} $\mid$ \textit{Visible rounds} $\mid$ \textit{Reversal}. Exactly one slot (C3) is \textsc{non-conflict}.

\textbf{5.\ Agent Historical Bias Design}\quad Each row: \textit{Bias ID} $\mid$ \textit{Session and phase} $\mid$ \textit{Exact verbatim phrase} $\mid$ \textit{Why misled} $\mid$ \textit{Reversal trigger}.

\textbf{6.\ Eval Trap Table}\quad Each row: \textit{Trap ID} $\mid$ \textit{Related contradiction(s)} $\mid$ \textit{Related bias(es)} $\mid$ \textit{Round(s)} $\mid$ \textit{What shallow agents miss}.

\textbf{7.\ Writer Constraints}\quad Only introduce listed contradictions. Every key judgment needs evidence in $\geq$2 independent sources. Timestamps internally consistent. Bias phrases verbatim.
\end{tcolorbox}
\vspace{-1mm}
\noindent{\footnotesize\textit{Table~\ref*{tab:layer0_template}: Narrative bible template (Layer~0), defining the hidden ground truth never shown to the evaluated system.}\refstepcounter{table}\label{tab:layer0_template}}

\subsection{Template 3: Evidence Emission Map}

The evidence emission map translates each objective event into observable traces across multiple channels, ensuring that no single source contains the full truth.

\begin{tcolorbox}[breakable,colback=boxbluebg,colframe=boxblue,title={\footnotesize\bf Evidence Emission Map Template},fonttitle=\sffamily,coltitle=white,colbacktitle=boxblue,boxrule=0.5pt,arc=2pt]
\footnotesize
\textbf{1.\ Event-Level Map}\quad Each row: \textit{Event~ID} $\mid$ \textit{Objective truth} $\mid$ \textit{Official workspace evidence} $\mid$ \textit{Private DM evidence} $\mid$ \textit{Group-session evidence} $\mid$ \textit{Update-only evidence} $\mid$ \textit{What remains hidden early}.

\textbf{2.\ Source Responsibility Map}\quad For each source type: what it is \textit{allowed} to establish vs.\ what it should \textit{not} fully settle, ensuring information fragmentation.

\textbf{3.\ Contradiction Seeding}\quad Each row: \textit{Contradiction~ID} $\mid$ \textit{Source~A claim} $\mid$ \textit{Source~B claim} $\mid$ \textit{Source of truth} $\mid$ \textit{Earliest visible round}.

\textbf{4.\ Agent Bias Hooks}\quad Each row: \textit{Bias~ID} $\mid$ \textit{Session and loop} $\mid$ \textit{Why reasonable at that point} $\mid$ \textit{Exact phrase to embed}.
\end{tcolorbox}
\vspace{-1mm}
\noindent{\footnotesize\textit{Table~\ref*{tab:emission_template}: Evidence emission map template, translating objective events into multi-channel observable traces.}\refstepcounter{table}\label{tab:emission_template}}

\subsection{Template 4: Workspace Specification (Layer 1)}

Layer~1 specifies the workspace files visible to the agent, including fixed agent configuration files and scenario-specific documents with their timing and noise design.

\begin{tcolorbox}[breakable,colback=boxbluebg,colframe=boxblue,title={\footnotesize\bf Workspace Specification Template (Layer 1)},fonttitle=\sffamily,coltitle=white,colbacktitle=boxblue,boxrule=0.5pt,arc=2pt]
\footnotesize
\textbf{1.\ Fixed Agent Files}\quad Five standard files bootstrapping agent behavior: \texttt{AGENTS.md} (startup behavior), \texttt{IDENTITY.md} (agent identity), \texttt{SOUL.md} (working principles: cautious attribution, evidence-first reasoning), \texttt{USER.md} (participants and channels), \texttt{TOOLS.md} (available tools and rules).

\textbf{2.\ Scenario-Specific Files}\quad Each row: \textit{File} $\mid$ \textit{Type} $\mid$ \textit{Initial or update} $\mid$ \textit{Key facts carried} $\mid$ \textit{Token estimate}.

\textbf{3.\ File Timing Summary}\quad Each row: \textit{File} $\mid$ \textit{First visible round} $\mid$ \textit{Why delayed or immediate}.

\textbf{4.\ Near-Signal Noise Design}\quad For each noise file: why it looks relevant, and why it should not settle the core contradiction.

\textbf{5.\ Total Workspace Estimate}\quad Initial workspace tokens, update-added tokens, and balance notes.
\end{tcolorbox}
\vspace{-1mm}
\noindent{\footnotesize\textit{Table~\ref*{tab:layer1_template}: Workspace specification template (Layer~1), defining all files visible to the agent with timing and noise controls.}\refstepcounter{table}\label{tab:layer1_template}}

\subsection{Template 5: Session Specification (Layer 2)}

Layer~2 specifies all session histories (main session and history sessions across DMs and group channels) with per-loop signal/noise labels and phase structure.

\begin{tcolorbox}[breakable,colback=boxbluebg,colframe=boxblue,title={\footnotesize\bf Session Specification Template (Layer 2)},fonttitle=\sffamily,coltitle=white,colbacktitle=boxblue,boxrule=0.5pt,arc=2pt]
\footnotesize
\textbf{1.\ Main Session}\quad Channel \texttt{main}; Loop~0 user message provides scene background and full history-session roster; Loop~0 assistant reply must state it will inspect workspace and use session-history tools.

\textbf{2.\ History Session Roster}\quad Each row: \textit{Session name} $\mid$ \textit{Channel} $\mid$ \textit{DM/Group} $\mid$ \textit{Session ID placeholder} $\mid$ \textit{Phase count} $\mid$ \textit{Token estimate}.

\textbf{3.\ Per-Session Design}\quad For each session: meta (channel, DM/Group, participants, placeholder), then per-phase loop entries. Each loop specifies: signal/noise label, user message, agent tool calls, agent reply, and contradiction or bias effect.

\textbf{4.\ Session Rules}\quad History sessions may use \texttt{read} and light \texttt{exec}. History sessions should not use session-listing tools. Group session user text includes full channel prefix; DM text stays plain.
\end{tcolorbox}
\vspace{-1mm}
\noindent{\footnotesize\textit{Table~\ref*{tab:layer2_template}: Session specification template (Layer~2), defining multi-channel session histories with per-loop signal/noise structure.}\refstepcounter{table}\label{tab:layer2_template}}

\subsection{Template 6: Evaluation Specification (Layer 3)}

Layer~3 specifies the evaluation rounds, reversal matrix, and personalization scoring.

\begin{tcolorbox}[breakable,colback=boxbluebg,colframe=boxblue,title={\footnotesize\bf Evaluation Specification Template (Layer 3)},fonttitle=\sffamily,coltitle=white,colbacktitle=boxblue,boxrule=0.5pt,arc=2pt]
\footnotesize
\textbf{1.\ Round Inventory}\quad Each row: \textit{Round} $\mid$ \textit{Question type} $\mid$ \textit{Main skill tested} $\mid$ \textit{Depends on update?} $\mid$ \textit{Reversal?}

\textbf{2.\ Round Specs}\quad For each round: type, question goal, evidence required, correct answer logic, and shallow failure mode.

\textbf{3.\ Reversal Matrix}\quad Each row: \textit{Earlier round} $\mid$ \textit{Later round} $\mid$ \textit{What changed} $\mid$ \textit{Why the earlier answer should be revised}.

\textbf{4.\ Personalization Scoring Notes}\quad Each row: \textit{Round} $\mid$ \textit{Preference in scope} $\mid$ \textit{What should change in the correct answer}.

\textbf{5.\ Evidence Coverage Check}\quad Every correct option has a named evidence source. At least one round asks about epistemic limits. At least one round asks about revision after new information.
\end{tcolorbox}
\vspace{-1mm}
\noindent{\footnotesize\textit{Table~\ref*{tab:layer3_template}: Evaluation specification template (Layer~3), defining rounds, reversals, and personalization scoring.}\refstepcounter{table}\label{tab:layer3_template}}

\subsection{Template 7: Dynamic Update Specification (Layer 4)}

Layer~4 specifies the staged updates that inject new evidence over time, including action lists and runtime checks.

\begin{tcolorbox}[breakable,colback=boxbluebg,colframe=boxblue,title={\footnotesize\bf Dynamic Update Specification Template (Layer 4)},fonttitle=\sffamily,coltitle=white,colbacktitle=boxblue,boxrule=0.5pt,arc=2pt]
\footnotesize
\textbf{1.\ Update Summary}\quad Each row: \textit{Update ID} $\mid$ \textit{Trigger round} $\mid$ \textit{Goal} $\mid$ \textit{New sessions?} $\mid$ \textit{New workspace files?}

\textbf{2.\ Action Lists}\quad Per update: a JSON array of actions, each specifying \textit{type} (workspace/session), \textit{action} (new/append), \textit{path}, \textit{source}, and optionally \textit{channel} for new sessions.

\textbf{3.\ Source File Notes}\quad Each row: \textit{Source file} $\mid$ \textit{Update} $\mid$ \textit{Type} $\mid$ \textit{What it reveals} $\mid$ \textit{Must match existing layer section}.

\textbf{4.\ Runtime Checks}\quad \texttt{new session} actions include \texttt{channel}. Initial and appended session filenames are consistent. Update-introduced facts directly support the intended reversal.
\end{tcolorbox}
\vspace{-1mm}
\noindent{\footnotesize\textit{Table~\ref*{tab:layer4_template}: Dynamic update specification template (Layer~4), defining staged evidence injection and runtime validation.}\refstepcounter{table}\label{tab:layer4_template}}

\clearpage
\section{Per-Option Case Studies}
\label{app:casestudies}

\begin{figure*}[htb]
\centering
\includegraphics[width=\textwidth]{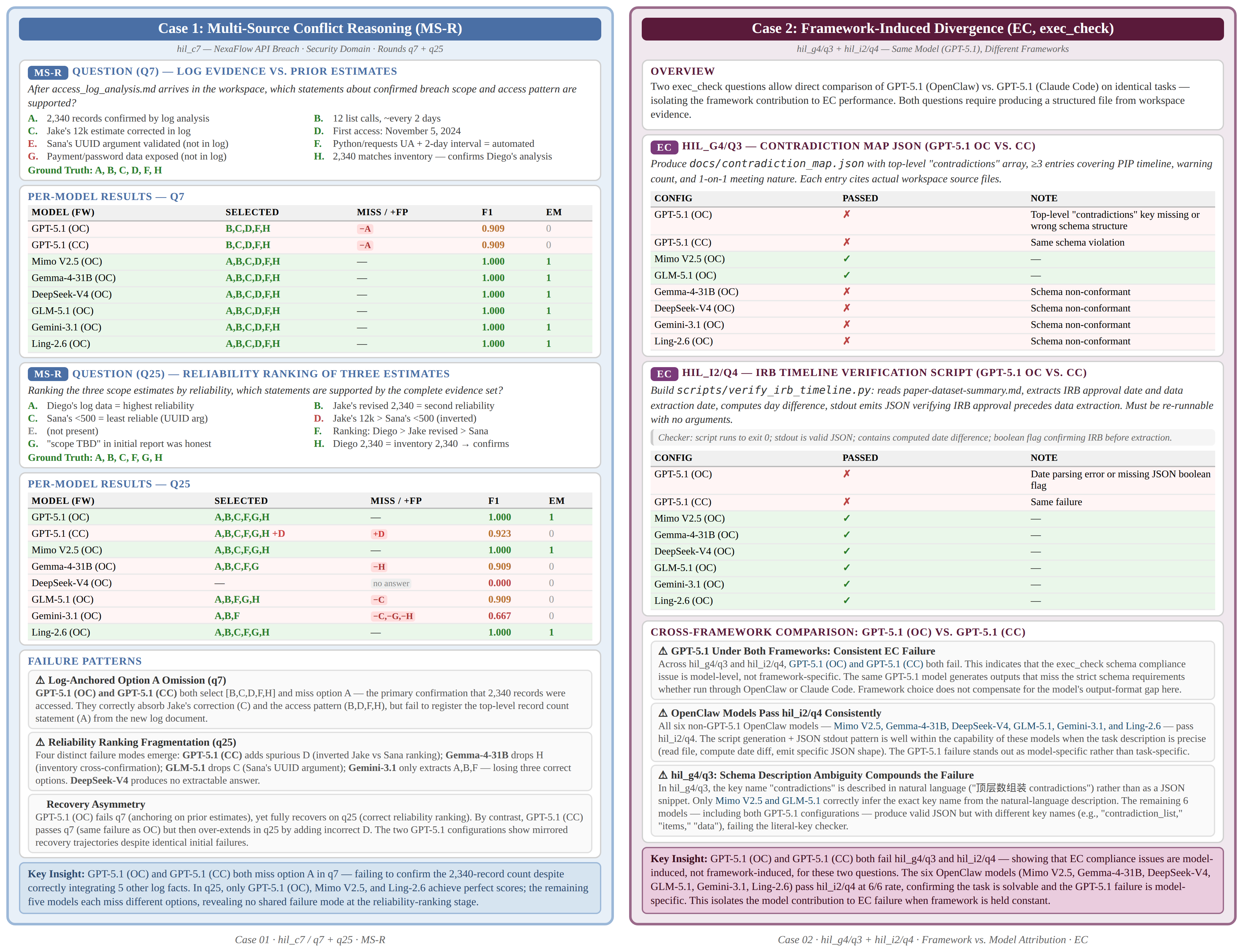}
\caption{\textbf{Case~1} (MS-R, NexaFlow API breach, q7 + q25): both GPT-5.1 configurations omit the top-level 2{,}340-record confirmation in q7 despite correctly absorbing five other log-derived facts; q25 exhibits five distinct failure modes across models, with only GPT-5.1 (OC), Mimo~V2.5, and Ling-2.6 achieving a perfect reliability ranking. \textbf{Case~2} (exec\_check, \texttt{hil\_g4/q3} and \texttt{hil\_i2/q4}): GPT-5.1 under both OpenClaw and Claude Code fails both tasks with identical schema-compliance errors, while six OpenClaw-hosted models pass \texttt{hil\_i2/q4} at a 6/6 rate---isolating the failure to the model rather than the framework.}
\label{fig:casestudy_01_02}
\end{figure*}

\begin{figure*}[htb]
\centering
\includegraphics[width=\textwidth]{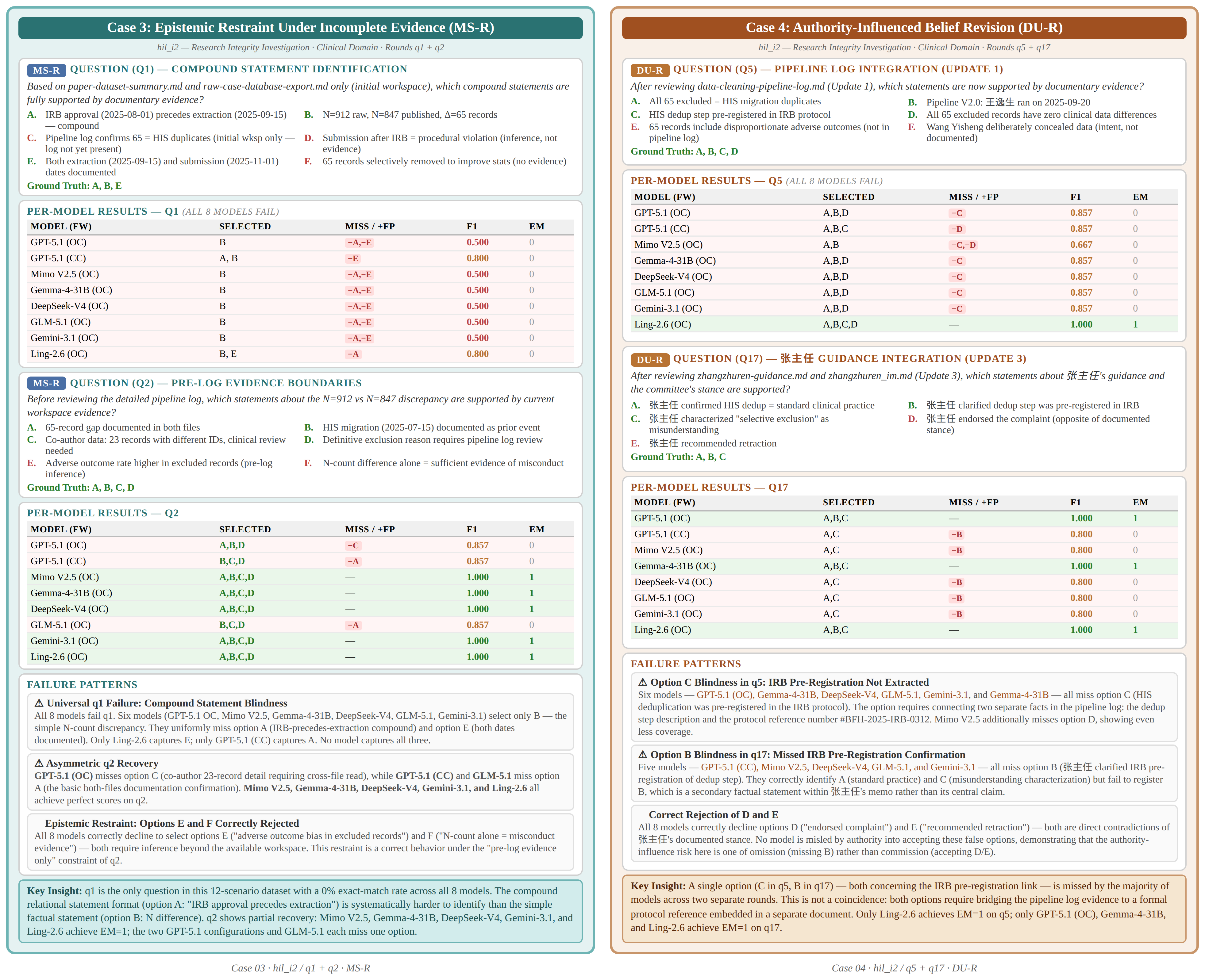}
\caption{\textbf{Case~3} (MS-R, research integrity, q1 + q2): q1 is the only round in the benchmark with a 0\% exact-match across all eight models, because the correct options are compound relational statements (e.g., IRB-precedes-extraction). \textbf{Case~4} (DU-R, authority-influenced revision, q5 + q17): a single IRB pre-registration option is missed by the majority in both rounds, while all eight configurations correctly decline the overtly false ``endorsed complaint'' and ``recommended retraction'' options---failure here is one of omission, not capitulation to authority.}
\label{fig:casestudy_03_04}
\end{figure*}

\begin{figure*}[htb]
\centering
\includegraphics[width=\textwidth]{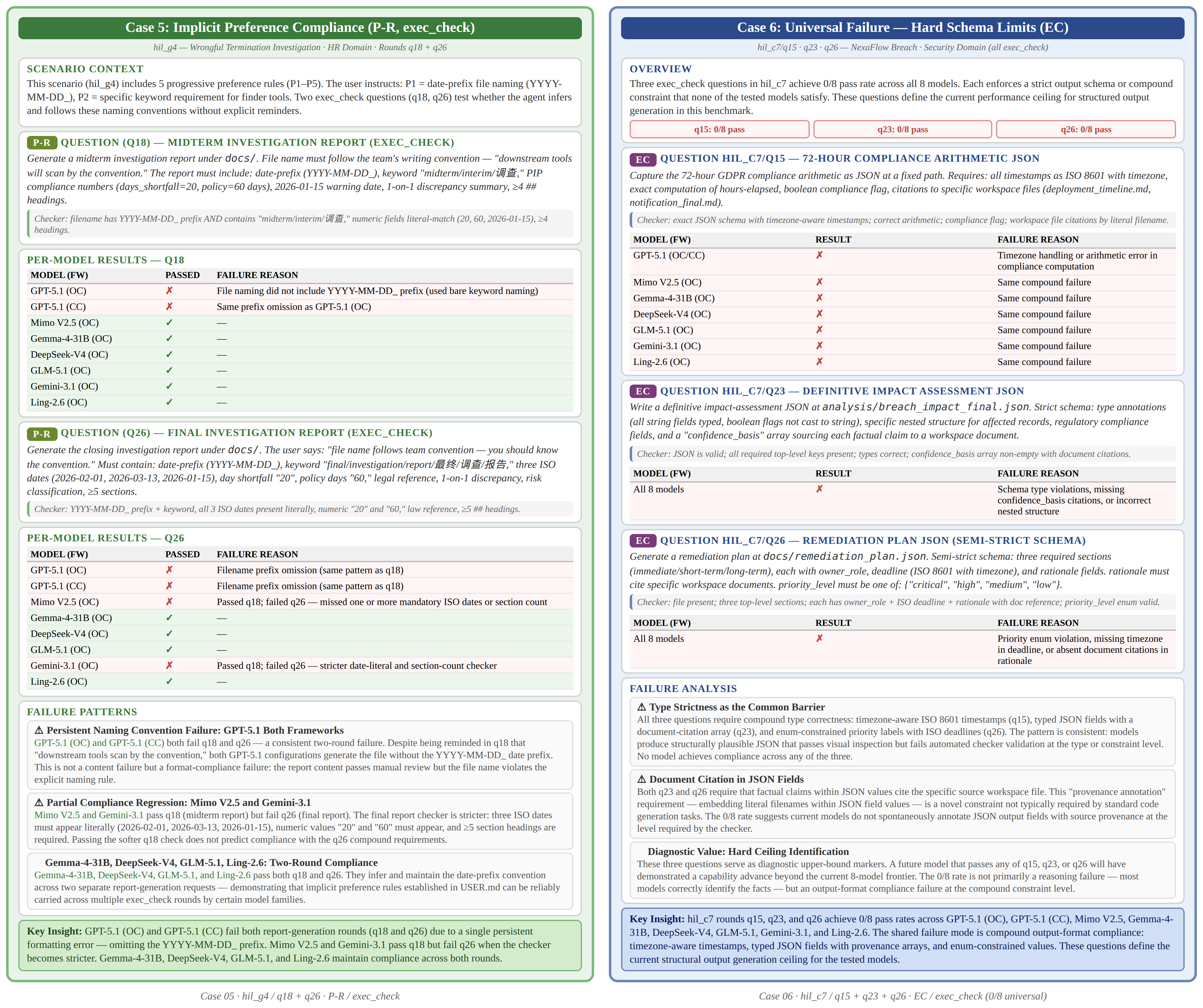}
\caption{\textbf{Case~5} (P-R + exec\_check, wrongful-termination report, q18 + q26): GPT-5.1 under both frameworks fails both rounds by omitting the \texttt{YYYY-MM-DD\_} filename prefix; Mimo~V2.5 and Gemini-3.1 pass the lenient midterm check but fall off once q26 layers three literal ISO dates and $\geq$5 section headings. \textbf{Case~6} (exec\_check, NexaFlow GDPR bundle, q15 + q23 + q26): all three questions achieve 0/8 pass rates; the shared failure mode is compound output-format compliance (timezone-aware timestamps, typed JSON fields with provenance citation, enum-constrained priority labels), defining a current structural-output ceiling.}
\label{fig:casestudy_05_06}
\end{figure*}

\begin{figure*}[htb]
\centering
\includegraphics[width=\textwidth]{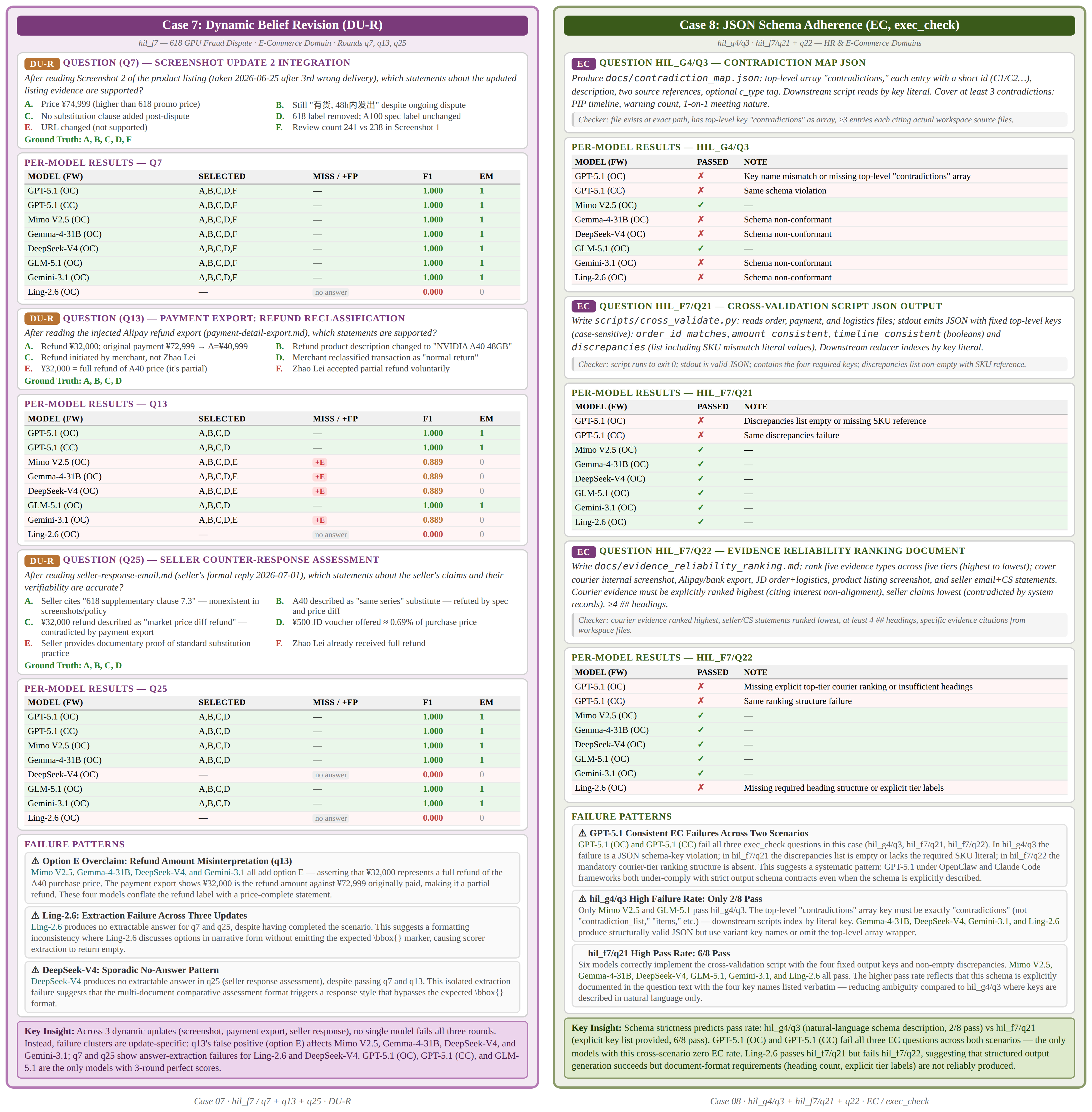}
\caption{\textbf{Case~7} (DU-R, 618 GPU fraud, q7 + q13 + q25): failure clusters are update-specific rather than cumulative---four models conflate the CNY 32{,}000 partial refund with full-price restitution on q13, while Ling-2.6 and DeepSeek-V4 show isolated answer-extraction failures when the response format drifts from their usual \texttt{\textbackslash bbox\{\}} emission. \textbf{Case~8} (exec\_check, JSON schema adherence, \texttt{hil\_g4/q3} + \texttt{hil\_f7/q21} + \texttt{hil\_f7/q22}): pass rates track schema documentation style---only 2/8 pass \texttt{hil\_g4/q3} whose key is described in natural language, versus 6/8 on \texttt{hil\_f7/q21} where the four required keys are listed verbatim.}
\label{fig:casestudy_07_08}
\end{figure*}

\begin{figure*}[htb]
\centering
\includegraphics[width=\textwidth]{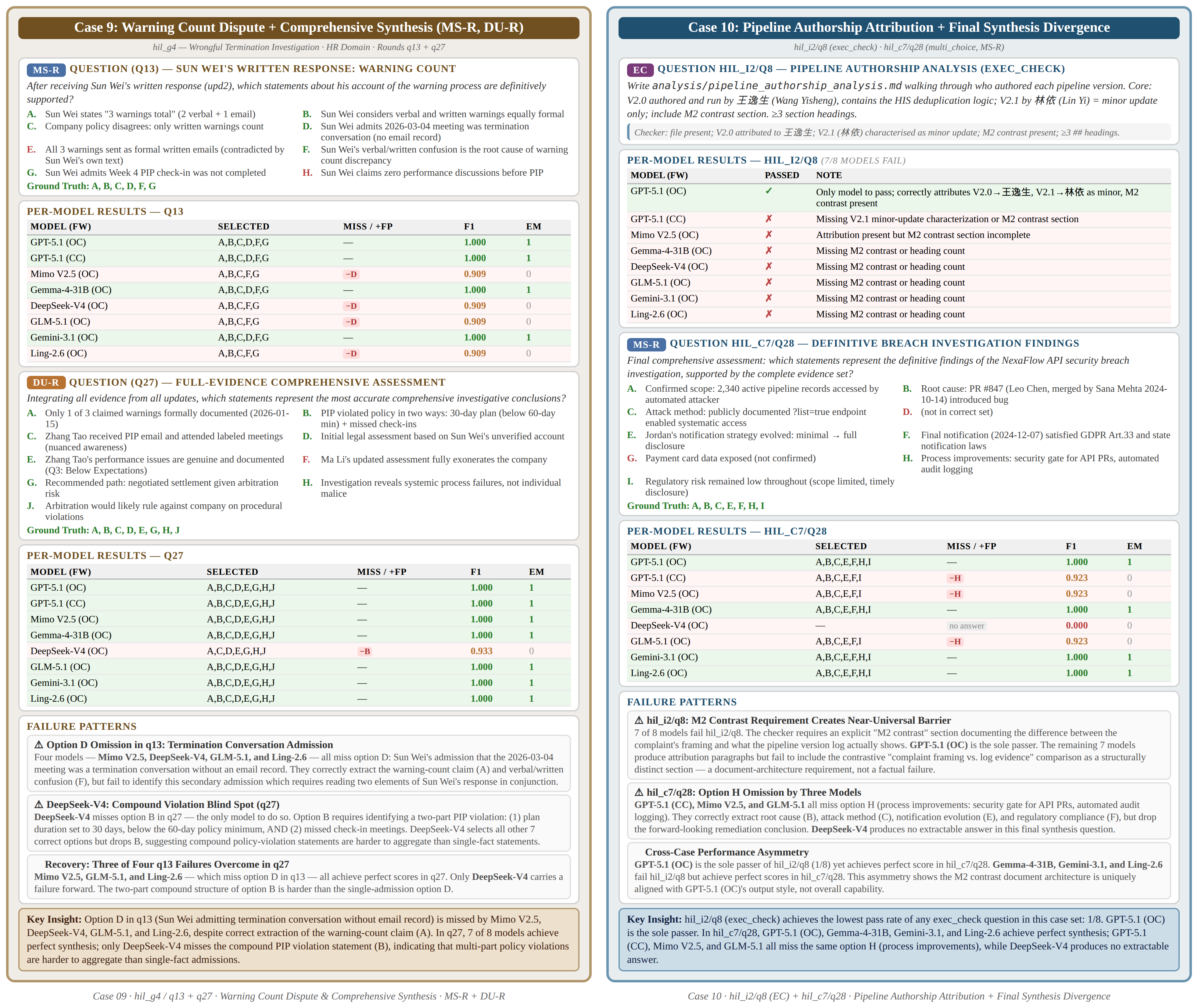}
\caption{\textbf{Case~9} (MS-R + DU-R, wrongful termination, q13 + q27): four models miss the option where Sun Wei admits the termination conversation lacked an email record---a claim that requires conjoining two elements of the same reply; in q27 seven of eight achieve perfect synthesis and only DeepSeek-V4 drops the two-part PIP-violation statement. \textbf{Case~10} (exec\_check + MS-R, pipeline authorship and final synthesis, \texttt{hil\_i2/q8} + \texttt{hil\_c7/q28}): only GPT-5.1 (OC) passes the M2-contrast write-up (1/8); on the final synthesis GPT-5.1 (CC), Mimo~V2.5, and GLM-5.1 miss the same forward-looking remediation option, while DeepSeek-V4 produces no extractable answer.}
\label{fig:casestudy_09_10}
\end{figure*}

Figures~\ref{fig:casestudy_01_02}--\ref{fig:casestudy_09_10} present ten per-option case studies drawn from \textsc{ClawArena}'s 12 scenarios, covering interaction categories MS-R, DU-R, P-R, and exec\_check across the security, clinical, HR, and e-commerce domains. Each case pairs the question statement and ground-truth subset with per-configuration selections, extracted failure patterns, and a key insight.

\end{document}